\documentclass[10pt,journal,compsoc]{IEEEtran}
\usepackage{amsmath,amssymb}
\usepackage{mathrsfs}
\usepackage{multirow}
\usepackage{hyperref}
\usepackage{color}
\usepackage{booktabs}
\newcommand{\hu}[1]{{\color{blue}{[#1]}}}

\newcommand{\tabincell}[2]{\begin{tabular}
{@{}#1@{}}#2
\end{tabular}}

\ifCLASSOPTIONcompsoc
  \usepackage[nocompress]{cite}
\else
  \usepackage{cite}
\fi

\ifCLASSINFOpdf
  \usepackage[pdftex]{graphicx}
\else
\fi
\begin{document}
%
\title{Convolutional Neural Networks with Gated Recurrent Connections}
%
%
%
%

\author{Jianfeng~Wang,
        Xiaolin~Hu,~\IEEEmembership{Senior Member,~IEEE}
\IEEEcompsocitemizethanks{\IEEEcompsocthanksitem Jianfeng Wang is with the Computer Science Department, University of Oxford,
 Oxfordshire, United Kindom.
E-mail: jianfeng.wang@cs.ox.ac.uk.
\IEEEcompsocthanksitem X. Hu is with the Institute for Artificial Intelligence, the State Key Laboratory of Intelligent Technology and
Systems, Beijing National Research Center for Information Science and Technology, and Department of Computer
Science and Technology, Tsinghua University, Beijing 100084, China.
E-mail: xlhu@tsinghua.edu.cn.}
}

\IEEEtitleabstractindextext{%
\begin{abstract}
The convolutional neural network (CNN) has become a basic model for solving many computer vision problems. In recent years, a new class of CNNs, recurrent convolution neural network (RCNN), inspired by abundant recurrent connections in the visual systems of animals, was proposed. The critical element of RCNN is the recurrent convolutional layer (RCL), which incorporates recurrent connections between neurons in the standard convolutional layer. With increasing number of recurrent computations, the receptive fields (RFs) of neurons in RCL expand unboundedly, which is inconsistent with biological facts. We propose to modulate the RFs of neurons by introducing gates to the recurrent connections.  The gates control the amount of context information inputting to the neurons and the neurons' RFs therefore become adaptive. The resulting layer is called gated recurrent convolution layer (GRCL). Multiple GRCLs constitute a deep model called gated RCNN (GRCNN). The GRCNN was evaluated on several computer vision tasks including object recognition, scene text recognition and object detection, and obtained much better results than the RCNN. In addition, when combined with other adaptive RF techniques, the GRCNN demonstrated competitive performance to the state-of-the-art models on benchmark datasets for these tasks. The codes are released at \href{https://github.com/Jianf-Wang/GRCNN}{https://github.com/Jianf-Wang/GRCNN}.
\end{abstract}

}

\maketitle

\IEEEdisplaynontitleabstractindextext

%
\IEEEpeerreviewmaketitle

\IEEEraisesectionheading{\section{Introduction}\label{sec:introduction}}

\IEEEPARstart{O}{ver} the past a few years, convolutional neural networks (CNNs), invented by LeCun {\it et al.} three decades ago \cite{lecun1989backpropagation}, have achieved great success in solving many computer vision problems. Without using handcrafted features, a CNN takes an image in pixels as input and extracts features automatically to accomplish a certain task, and the entire process can be trained end-to-end. Different CNN architectures have been proposed \cite{Krizhevsky2012ImageNet, Simonyan2014Very, He2016Deep, Huang2017Densely, Szegedy2014Going, xie2016aggregated} and the performance is getting better and better.

\begin{figure}
\includegraphics[width=\linewidth]{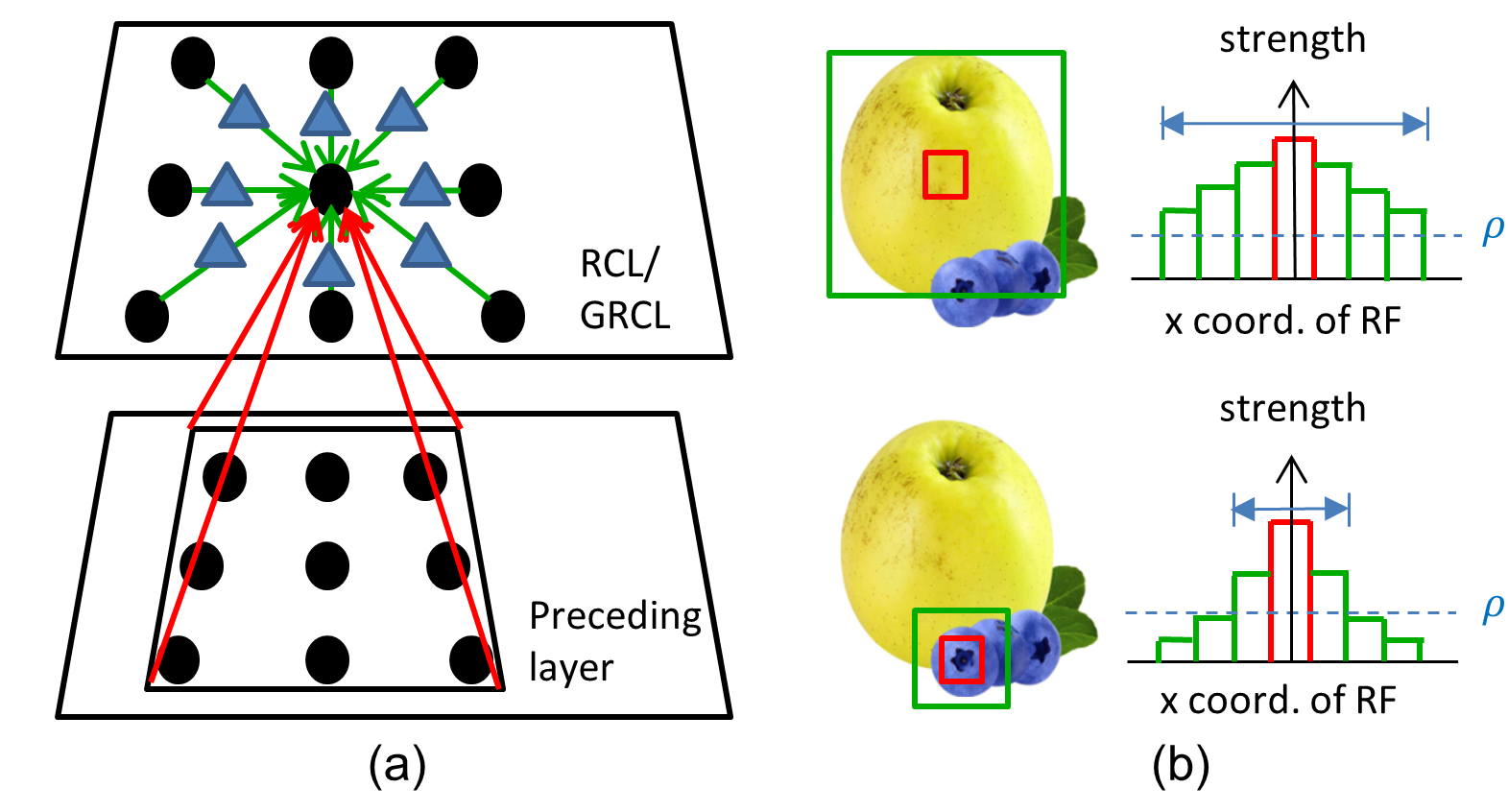}
\caption{Illustration of the RCL and GRCL. (a) Structures of the RCL and GRCL. The black dots denote neurons. The red and green
arrows denote the feedforward and recurrent connections, respectively. With the gates denoted by triangles operating on the
recurrent connections, the figure illustrates a GRCL; otherwise, it illustrates an RCL.
(b) Desired RFs in the GRCL in recognizing the pear and blueberry. \emph{Left:} The red squares and green squares denote the classical RF and the desired non-classical RF of a neuron. \emph{Right:} Desired strength of the non-classical RF. For clarity, only the x coordinate of RF is shown. The threshold $\rho$ determines the size of the
effective RF, {\it i.e.}, only regions in the RF where the strength is larger than $\rho$
are counted.}
\label{fig:motivation}
\end{figure}

In history, CNN has got many benefits from neuroscience.  For instance, the two standard operations in CNN, convolution and pooling, can date back to Hubel and Wiesel's discovery of simple cells and complex cells in the primary visual cortex (Vl) of cats \cite{Hubel1959Receptive, hubel1962receptive}. A predecessor of CNN is Neocognitron \cite{Fukushima1980Neocognitron} which also uses the two operations. Several years ago, inspired by abundant recurrent synapses within the same visual area such as V1 \cite{Dayan2001Theoretical}, a recurrent CNN (RCNN) \cite{Liang2015Recurrent, Ming2015Recurrent} was invented, and the main idea is to incorporate recurrent connections between neurons within the same convolutional layer. The resulting layer is called recurrent convolutional layer (RCL). Its structure is illustrated in Fig. \ref{fig:motivation}a.
A neuron receives inputs from its neighboring neurons in the same layer besides the neurons in the
preceding layer. For clarity, only recurrent connections in the
same feature map are plotted, but in practice every neuron
receives recurrent inputs from all feature maps in the current
layer. Similarly, every neuron receives feedforward inputs
from all feature maps in the preceding layer.
Due to the recurrent connections, object recognition becomes a dynamic process though the input is static. Since neurons in the same layer can exchange information, the receptive field  (RF) of every neuron, defined as the region in the input which can influence the activity of the neuron, will become larger than that of the neuron in a conventional convolutional layer (Fig. \ref{fig:motivation}). This agrees with neuroscience findings to some extent, as a biological neuron's activity can be modulated, either suppressed or facilitated, by stimuli outside the neuron's classical RF \cite{nelson1978orientation, jones2002spatial, cavanaugh2002nature}. In this sense, the RF induced by the feedforward connections in Fig. \ref{fig:motivation}a, {\it i.e.,} the red square projected to the input space, can be called {\it classical RF} while the RF induced by both feedforward and recurrent connections can be called {\it non-classical RF}. In what follows, if not specified, the RF in a model with recurrent connections refers to  the non-classical RF.

However, there is no restriction on this RF expanding process in the RCL. With a large number of recurrent iterations, all neurons in an RCL would ``see'' the whole input image and the contents captured in different iterations are equally fed to the neuron, which is neither consistent with biological facts nor advantageous for visual perception. Intuitively, to recognize the pear and the blueberry in Fig.~\ref{fig:motivation}b, it is better to use different RF sizes. However, in the RCL, the RF sizes of neurons increase with recurrent iterations regardless of the input. We need a mechanism to automatically control the sizes of RFs, or at least the strengths at different locations inside the RFs, to better ``understand'' the input.


Towards this goal, we propose to add gates onto the recurrent connections to modulate the RFs of neurons in the RCL, denoted by the triangles in Fig. \ref{fig:motivation}a. The resulting module is called gated RCL (GRCL). The outputs of the gates are between 0 and 1, and the value is determined by the input to the GRCL, as well as the states of the neurons in the previous recurrent iteration. Therefore the outputs of the gates in different iterations could be different. The recurrent contribution to the activities of neurons is multiplied with the corresponding outputs of the gates before adding to the feedforward contribution. If the gates always output 1 during recurrent iterations, the GRCL degenerates to the RCL.

It is expected that the gates could learn to modulate the RFs such that the neurons focus on regions with appropriate sizes for different inputs. Fig. \ref{fig:motivation}b \emph{left} shows desired RFs (green squares) for recognizing the pear and the blueberry. This goal can be achieved if all gates in the former case are open while some gates in later iterations in the latter case are closed (output 0). It can be partially achieved if all gates are open in the two cases but the gates output smaller values in the latter case than in the former case, because such a setting makes the recurrent contributions diminish along with iterations faster in the latter case. Fig. \ref{fig:motivation}b \emph{right} illustrates the desired strengths of the RFs in the two cases. Every green rectangle denotes the expansion of the RF in one recurrent iteration. Here, uniform feedforward kernels and recurrent kernels are assumed. That is why the strengths are step functions instead of smooth curve functions.  If a threshold $\rho$  (dashed blue line) is used to determine the size of the effective RF, the size of the effective RF would be smaller in the latter case.

The GRCL is a basic building block, and multiple GRCLs can be stacked into a deep network called {\it gated recurrent CNN} or GRCNN. This paper presents the implementation and experiments of GRCNN.


Some preliminary results have been presented at a conference \cite{jianfeng2017deep}. The present work extends the previous work in several aspects. First, we integrate some advanced techniques developed in recent years into the model to further improve its performance. Second, to determine the states of the neurons in the current iteration, we integrate recurrent contributions in all previous iterations instead of only the last iteration.
Third, we allow the recurrent weights to be different in different recurrent iterations. This relaxation allows us to use more parameters for fair comparison with the state-of-the-art models on certain computer vision tasks. In other words, some models presented in the paper are actually not {\it recurrent} networks. 
In this case, the recurrent connections become feedforward connections. The function of the gates on them will be elaborated in later sections.
Fourth, in the previous work, we only evaluated the model on the optical character recognition (OCR) task, but in this work we in addition evaluated the model on object recognition and object detection tasks. Extensive experiments showed that GRCNN obtained very good results in these tasks.

The remaining contents of the paper are organized as follows. In Section \ref{sec:related}, some closely related works are introduced. In section \ref{sec:model}, the architectures of RCNN and GRCNN are described. Sections \ref{sec:object-recognition}, \ref{sec:OCR} and \ref{sec:detection} present experimental results on object recognition, OCR and object detection, respectively. Section \ref{sec:conclusion} concludes the paper.

\section{Related Work}\label{sec:related}

\subsection{CNNs with Skip-Layer Connections}


Here we briefly review CNNs with skip-layer or shortcut connections because the proposed model can be viewed as such a model when unfolded through time.

The highway network \cite{Srivastava2015Training} is one of the earliest CNNs with skip-layer connections. The main idea is to add a bypassing path to the convolutional layer such that the output of the layer has two parts: one nonlinear transformation part (conventional convolution and nonlinear activation) and one bypassing part (the input of the layer). The two parts are modulated by a gate, usually implemented as the logistic sigmoid function whose output is between 0 and 1. When the gate on the bypassing path is 1, the input is directly added to the output of the layer without attenuation. In contrast, the residual network (ResNet) \cite{He2016Deep} advocates bypassing paths in CNN without gates. The argument is that without input attenuation, the convolutional layer learns the residual between the ideal mapping and the identity mapping from input to output. The strategy turns out to be very effective and ResNet has been applied to numerous applications. In addition, many improved versions have been proposed. For example, a more efficient model ResNeXt \cite{xie2016aggregated} is obtained by mainly substituting the conventional convolution in ResNet with the grouped convolution.
The densely connected network (DenseNet) \cite{Huang2017Densely} consists of multiple blocks and in each block every layer has direct paths from all layers below (this is the origin of its name) besides the nonlinear transformation path. In contrast to the highway network and ResNet which integrate two paths by summing their outputs, DenseNet integrates multiple paths by the concatenation operation. This modification ensures that the upper layers in a block do not lose the information of the input to the block. The squeeze-and-excitation network (SENet) \cite{Hu2017Squeeze} proposes to scale the feature maps of a convolutional layer using a squeeze-and-excitation operation. The squeeze-and-excitation path and the direct path between the original convolutional layer are integrated by feature map-wise product. An extension of SENet is the Gather-Excite network (GENet) \cite{hu2018gather}, which uses a different product operation for integrating the two paths.


\subsection{CNNs with Adaptive Receptive Fields}
There are a few works investigating adaptive RF in neural networks. A typical model refers to the Deformable CNN \cite{dai2017deformable}. Unlike the traditional CNN, the Deformable CNN has a changeable grid in the sense that each point in it can be moved by a learnable offset, and then the convolution filters operate on these moved grid points, resulting in the deformable convolution. The objective is to encourage the new convolutional block to learn features from the best location in the previous layer. The Saliency Network \cite{recasens2018learning} performs input distortions to ``zoom-in'' salient regions, which resembles deformable RF on input images. In the active convolution \cite{jeon2017active}, the RF does not have a fixed shape and can take diverse forms for convolutions. Inspired by the human visual system, a novel module is proposed to be integrated into CNN, named Receptive Field Block \cite{liu2018receptive}. It utilizes multi-branch pooling with varying kernels corresponding to RFs of different sizes, and then uses dilated convolution to control their eccentricities, and reshapes them to generate final representation. 
 Self-attention is considered to be an alternative to the convolutional layer \cite{cordonnier2020relationship}. The multi-head self-attention layer can also implement a dynamic receptive field, since each head can attend a value at any pixel shift and form different grid patterns. 
The Selective Kernel Network (SKNet) \cite{Xiang2019SKNet} consists of several "selective kernel" (SK) convolutions. The SK convolution is also inspired by the adaptive RF sizes of neurons, which is implemented by three operators, i.e., \textit{Split}, \textit{Fuse} and \textit{Select}. Our method implements adaptive RFs in a different way, {\it i.e.}, using gates to control the strength of different points in RFs of the RCNN \cite{Liang2015Recurrent}.

\subsection{Recurrent Neural Networks}
RNNs are often used to model sequential data such as handwriting recognition \cite{Graves2009A} and speech recognition \cite{Graves2013Speech}. Simple RNN has a notorious property, namely ``gradient vanishing'', which makes it difficult to train the RNN. To circumvent this drawback, the long short term memory (LSTM) \cite{Hochreiter1997Long} and the gated recurrent unit (GRU) \cite{Cho2014Learning} are proposed. Both LSTM and GRU adopt gates to control the information flow and determine the memory that should be kept.
%

There are not many RNNs proposed to process images, because it is not apparent how the build-in sequential process of an RNN could be used to process static images. In 2015, a recurrent CNN (RCNN) was proposed to perform object recognition \cite{Liang2015Recurrent}. The main idea is to introduce recurrent connections between neighboring neurons in the same convolutional layer, and the resulting layer is called RCL. Then the neurons' activities are no longer static because the neurons interact with others at each recurrent iteration. This is consistent with biological neurons whose firing rates are not constants over time given a stimulus \cite{kandel2000principles}. The model was shown to be more robust to occlusions in object recognition than feedforward models \cite{spoerer2017recurrent}. It was extended to a multi-scale version for scene labeling \cite{Ming2015Recurrent}. It was also used in other applications including speech recognition  \cite{Zhao2017Recurrent}, action recognition \cite{Wang2016Deep} and OCR \cite{Lee2016Recursive}.

When unfolded through time, RCNN becomes a feedforward network with many skip-layer connections. In the unfolded RCL, such skip-layer connections represent identity mappings from the input to every layer above. Every layer integrates the convolutional path and the identity mapping path using summation. GRCNN differs from RCNN by introducing gates to control information flow in the convolutional path, and it differs from the highway network \cite{Srivastava2015Training} by not using gates on the identity mapping path.

Another method to integrate RNN and CNN is to introduce recurrent connections between different layers rather than within the same layer. The convolutional deep belief network (CDBN) \cite{Lee+etal09:convDBN} is an earlier attempt. Every pair of adjacent layers has both bottom-up and top-down connections. Instead of using recurrent connections between adjacent layers, one can also use recurrent connections from the top layer to the bottom layer of a deep model \cite{Pinheiro2014Recurrent}.

\section{Model}\label{sec:model}

Since the proposed model is based on the RCNN model \cite{Liang2015Recurrent, Ming2015Recurrent}, for better understanding, we first briefly introduce the architecture of RCNN, then describe the proposed GRCNN in detail. As the RCNN and the GRCNN only replace the convolutional layer in conventional CNN with the RCL and the GRCL, respectively, we mainly describe the RCL and the GRCL. Note that the RCL described below slightly differs from its original version \cite{Liang2015Recurrent, Ming2015Recurrent} as we incorporate some later proposed techniques into it.

\subsection{Recurrent Convolutional Layer and RCNN}\label{sec:RCNN}
The RCL \cite{Liang2015Recurrent, Ming2015Recurrent} is a module with recurrent connections in the convolutional layer (Fig. \ref{fig:motivation}a without the gates). The states of the neurons $x(t)$ evolve as follows
\begin{align}\label{eqn:RCL}
 &x(t)=f( w^\text{F}*u(t) + w^\text{R}*x(t-1))
\end{align}
for $t\ge 0$, where ``*'' denotes convolution, $u(t)$ and $x(t-1)$ denote the feed-forward input and recurrent input to the RCL, respectively, $w^\text{F}$ and $w^\text{R}$ denote feed-forward weight and recurrent weights, respectively, and $f(\cdot)$ denotes the activation function (usually the ReLU function). We stress that $t$ indexes the processing steps in the model and it does not suggest that the image is moving or in other dynamic modes. Throughout the paper, we assume $x(t)=0$ for $t<0$. Therefore, when $t=0$ in \eqref{eqn:RCL}, we only have the first term in $f(\cdot)$.

Because any neuron's activity is influenced by the neighboring neurons' activities through recurrent weights $w^\text{R}$, as the recurrent computation continues, the neuron would ``see'' larger and larger area in the input. In other words, the sizes of RFs of neurons would increase unboundedly.

\begin{figure}
\centerline{\includegraphics[width=\linewidth]{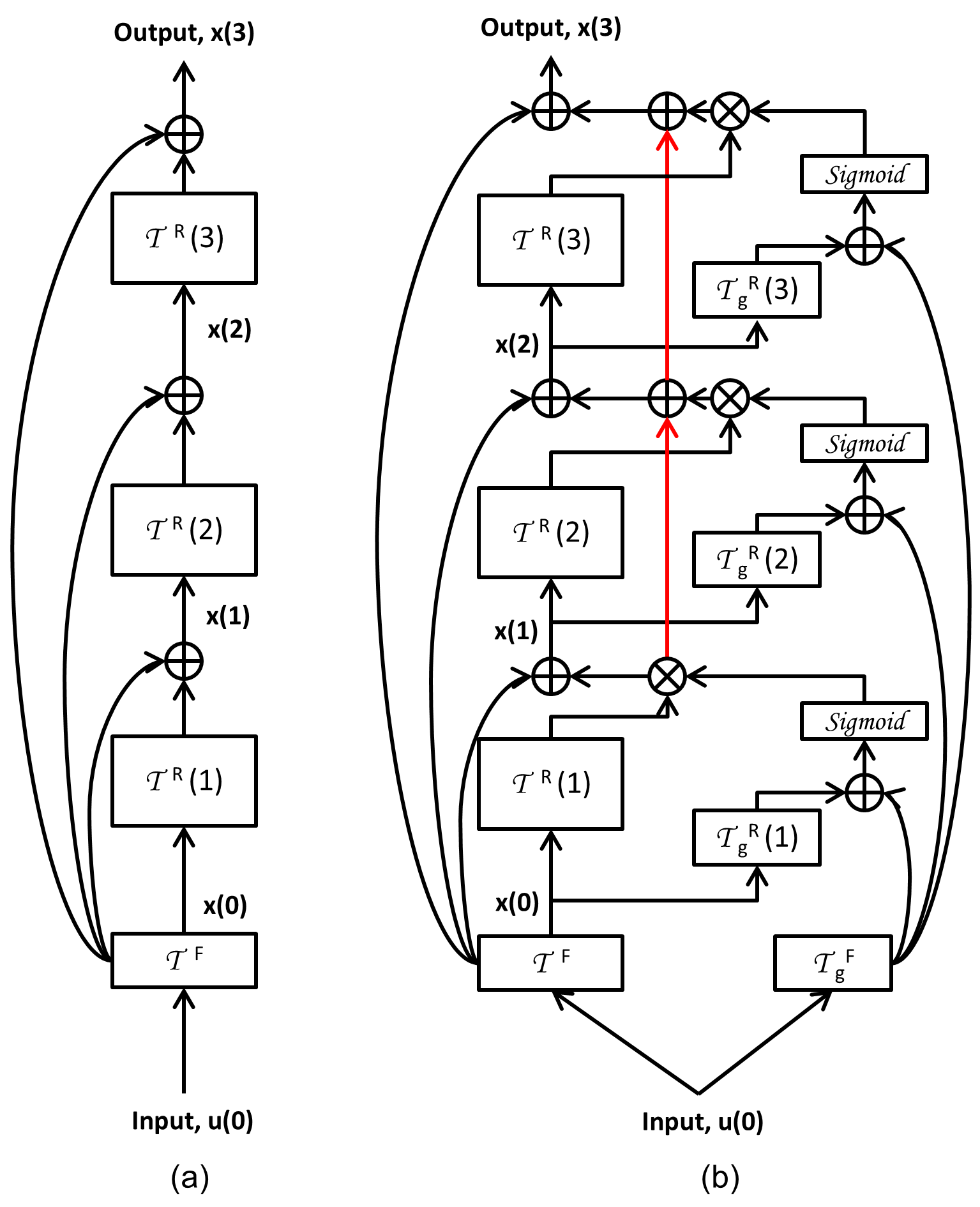}}
\caption{Time-unfolded version of the RCL (a) and the GRCL (b) with $T=3$. The diagram in (b) without the red arrows corresponds to the model presented in the preliminary work \cite{jianfeng2017deep}.  }
\label{fig:GRCL}
\end{figure}

To learn the parameters, one needs to unfold the recurrent computation through time in the RCL to construct a feedforward architecture where the weights between layers are identical, {\it i.e.}, all are $w^\text{R}$. The number of recurrent iterations is a hyper-parameter and task-dependent. In addition, the input $u(t)$ is present in each recurrent iteration, {\it i.e.}, in each layer of the time-unfolded architecture (Fig. \ref{fig:GRCL}a). Note that unfolding an RCNN with multiple RCLs has different methods, which lead to different feedforward architectures \cite{Ming2015Recurrent}. Here, we adopt the method in which RCLs are unfolded one by one in the bottom-up direction. In other words, we unfold an RCL through time and use the output in the last iteration as input to the layer above, while the outputs in other iterations are not directly inputted to the layer above. In this method, $u(t)$ does not change over time, and in what follows, we will drop $t$ in the input.

To improve the performance of the RCNN, two modifications are made in this work. First, we incorporate recently proposed batch normalization (BN) \cite{Ioffe2015Batch} and pre-activation techniques \cite{He2016Identity}. The former is proposed to reduce the so-called {\it internal covariate shift} (but a recent study suggests that its actual effect is to smooth the landscape of the loss function \cite{santurkar2018does}). The latter advocates to apply BN \cite{Ioffe2015Batch} and ReLU before each convolution, instead of ``Conv-BN-ReLU'' which is often used in prior works. Both techniques have been proved very effective in deep neural networks. Then the states of the neurons evolve as follows
\begin{align}\label{eqn:RCL1}
 &x(t)= w^\text{F}*f(\text{BN}(u)) + w^\text{R}(t-1)*f(\text{BN}(x(t-1))).
\end{align}
In the original proposal \cite{Liang2015Recurrent, Ming2015Recurrent}, $w^\text{R}(t)$ was identical for different $t$.

Second, as explained in Introduction, we remove the weight sharing constraint in the time-unfolded architecture. In other words, different $w^\text{R}(t)$ are used for different $t$. This is to use more parameters for fair comparison with other models on certain tasks. However, no matter whether $w^\text{R}(t)$ is shared or not across time steps, the statistics and the parameters used for the affine transformation in BN are not shared throughout this paper since $x(t-1)$ is not identical with different $t$. Note that $w^\text{F}$ does not depend on $t$ and it still remains the same across $t$.
From the perspective of time-unfolded  network, the RF size of neurons always increases along the hierarchy, say, higher layer neurons have larger RFs, no matter $w^\text{R}$ is shared or not. The only difference is that in the weight sharing case, the increase can be viewed as a dynamic process of the same set of neurons, while in the weight unsharing case we do not have this view.

RCL \eqref{eqn:RCL1} can be rewritten in a more general form,
\begin{align}\label{eqn:RCL2}
 &x(t)= \mathcal{T}^\text{F}(u; w^\text{F})+\mathcal{T}^\text{R}(x(t-1);w^\text{R}(t-1)),
\end{align}
where $\mathcal{T}^\text{F}$ and $\mathcal{T}^\text{R}$ denote the feed-forward transformation with parameters $w^\text{F}$ and recurrent transformation with parameters $w^\text{R}$ using the pre-activation technique, respectively. For simplicity, BN and ReLU are omitted in equation~(\ref{eqn:RCL2}), and
$\mathcal{T}^\text{F}$ is still implemented by a simple "BN-ReLU-Conv" layer throughout the paper. As for $\mathcal{T}^\text{R}$, it can be implemented by different blocks, such as a simple ``BN-ReLU-Conv'', a general 3-layer bottleneck block \cite{He2016Deep, He2016Identity}, or a 3-layer bottleneck block with SK convolution \cite{Xiang2019SKNet}.

\begin{table*}
\centering
\caption{Test Error Rates of RCNN and GRCNN on CIFAR-10 (\%). }
\footnotesize
\label{tab:compare-cifar10}
\begin{tabular}{ccccccccc}
\toprule
\multirow{2}{*}{Iterations} & \multirow{2}{*}{Feature maps}  & \multicolumn{3}{c}{Tied weights}&&\multicolumn{3}{c}{Untied weights}\\
\cline{3-5}\cline{7-9}
&&RCNN &Original GRCNN& Improved GRCNN && RCNN & Original GRCNN & Improved GRCNN \\
\toprule
(2,2,2) & (128,128,128)&5.81 $\pm$ 0.13 & 5.92 $\pm$ 0.10&5.83 $\pm$ 0.13 && 5.55 $\pm$ 0.11 &5.71 $\pm$ 0.11 & 5.65 $\pm$ 0.04\\

(3,3,3) &(128,128,128)&5.65 $\pm$ 0.09 &5.80 $\pm$ 0.12 & 5.76 $\pm$ 0.07 &&  5.39 $\pm$ 0.05 & 5.53 $\pm$ 0.12 &5.46 $\pm$ 0.10\\

(4,4,4) &(128,128,128)& 5.76 $\pm$ 0.06& 5.69 $\pm$ 0.11& 5.65 $\pm$ 0.07&& 5.51 $\pm$ 0.14 & 5.30 $\pm$ 0.13 & 5.24 $\pm$ 0.07 \\

(5,5,5) & (128,128,128)&5.88 $\pm$ 0.11 & 5.57 $\pm$ 0.08&5.49 $\pm$ 0.12 && 5.62 $\pm$ 0.09& 5.23 $\pm$ 0.09  & 5.16 $\pm$ 0.10 \\
\bottomrule
\end{tabular}
\end{table*}

\subsection{Original GRCL and GRCNN}\label{sec:originalGRCNN}
In an RCL, the RF size of neurons increases unboundedly with recurrent computation. Our primary goal is to introduce a mechanism to alleviate this problem. The basic idea is to modulate the recurrent term by a gate before adding it to the input term, and the resulting module is called gated RCL (GRCL). Integrated with the pre-activation technique, the original GRCL in our previous work \cite{jianfeng2017deep} can be described as follows (Fig. \ref{fig:GRCL}b, without the red arrows).
\begin{align}\label{eqn:GRCL-nips}
 &x(t)= \mathcal{T}^\text{F}(u; w^\text{F})+G(t)\odot\mathcal{T}^\text{R}(x(t-1);w^\text{R}(t-1))
\end{align}
for $t\ge0$, where $\odot$ denotes element-wise multiplication, and $G(t)$ is a gate whose outputs have the same dimension as $\mathcal{T}^\text{R}(x(t-1);w^\text{R}(t-1))$ and $x(t)$,
\begin{equation}\label{eqn:gate}
G(t)= \sigma( \mathcal{T}_\text{g}^\text{F}(u; w^\text{F}_\text{g})+\mathcal{T}_\text{g}^\text{R}(x(t-1); w^\text{R}_\text{g}(t-1)) )
\end{equation}
for $t\ge 0$ with $\sigma$ being the logistic sigmoid function: $\sigma(x)=1/(1+\exp(-x))$. In the equation, $\mathcal{T}^\text{F}_\text{g}$ and $\mathcal{T}^\text{R}_\text{g}$ denote the feed-forward transformation and recurrent transformation using the pre-activation technique, respectively, for determining the output of the gate, and they have their own parameters $w^\text{F}_\text{g}$ and $w^\text{R}_{\text{g}}(t)$.
Note that $w^\text{R}_{\text{g}}(t)$ can be shared across $t$ or not. For simplicity, $w^\text{F}_\text{g}$ and $w^\text{R}_{\text{g}}(t)$ are implemented by a simple ``BN-ReLU-Conv'' separately throughout  the  paper. The convolutional layers in $w^\text{F}_\text{g}$ and $w^\text{R}_{\text{g}}(t)$ are a set of $1\times1$ convolutional filters in order to make the dimensions of the transformed $u$ and $x(t-1)$ consistent with $G(t)$.

If tied weights across layers are used (as in \cite{jianfeng2017deep}), \eqref{eqn:GRCL-nips} and \eqref{eqn:gate} describe the dynamics of a set of neurons. As explained before, the expansion of the RFs of neurons is incurred by recurrent computation. By comparing \eqref{eqn:RCL2} and \eqref{eqn:GRCL-nips}, we can see that the gate $G(t)$ weakens the recurrent contribution, which comes from a larger area in the input than from the RF in the previous recurrent iteration, because $G(t)<1$.
If $G(t)\ne 0$ and a threshold is used to delineate the effective RF, the size of the effective RF will depend on the value of $G(t)$ (Fig. \ref{fig:motivation}b).
In the case of untied weights, the effect of $G(t)$ is to modulate the RFs of neurons in different layers, allowing them to be adjusted adaptively according to the input. It should be noted that, for an untied GRCL, the feedforward weights are still shared across the time steps.

Multiple GRCLs together with other types of layers such as pooling layers can be stacked into a deep model. Hereafter, a CNN that contains a GRCL described in \eqref{eqn:GRCL-nips} is called an original GRCNN.

\subsection{Improved GRCL and GRCNN}\label{sec:revisedGRCNN}
However, the introduction of gates causes a new problem: the recurrent contribution to the states of the neurons could be very small. According to \eqref{eqn:GRCL-nips}, only the recurrent contribution in the last iteration is explicitly added to the feedforward contribution, and the result is propagated to the next layer. In one extreme case, all gates in the last iteration output zeros, then only the feedforward contribution is propagated to the next layer, which is clearly not what we want. Even if none of the gates outputs zero, the recurrent contributions in previous iterations are likely to decay quickly. This is mainly due to the recursive computation of gates \eqref{eqn:GRCL-nips} whose outputs fall into an interval ranging from 0 to 1.

To circumvent this deficiency, we propose to explicitly add recurrent contributions in all previous iterations to the feedforward contribution:
\begin{equation}\label{eqn:GRCL}
x(t)= \mathcal{T}^\text{F}(u; w^\text{F})+\sum_{n=1}^{t} {G(n)\odot\mathcal{T}^\text{R}(x(n-1); w^\text{R}(n-1))}
\end{equation}
for $t\ge1$, where $G(n)$ is defined in \eqref{eqn:gate}. For $t=0$, $x(t)= \mathcal{T}^\text{F}(u; w^\text{F})$. If the gates are closed at certain time steps, the recurrent information computed at other time steps still contributes to the state at the current time step $t$.  It is easy to verify that this modification corresponds to adding a few connections to the structure of the original GRCL (Fig.~\ref{fig:GRCL}b). Moreover, to save the number of parameters, the convolutional layer in $w^\text{F}$, $w^\text{F}_\text{g}$ and $w^\text{R}_{\text{g}}(t)$ are implemented by the grouped convolution. This new version of GRCL is called {\it improved GRCL}.

Hereafter, a CNN that contains an improved GRCL is called an {\it improved GRCNN}.

\section{Object Recognition}\label{sec:object-recognition}
The first application of the proposed GRCNN is object recognition. We first designed models for three small datasets and performed detailed analyses on one of them, then designed models for a large dataset.

\subsection{Datasets}
{\bf CIFAR:} We conducted experiments on two CIFAR datasets \cite{Krizhevsky2009Learning} which consist of colored natural images with $32 \times 32$ pixels. The images in the two datasets are the same but the labels are different. The CIFAR-10 dataset has 10 classes and the CIFAR-100 dataset has 100 categories. The training and test sets have 50,000 images and 10,000 images, respectively.

{\bf SVHN:} The street view house numbers (SVHN) dataset \cite{Netzer2011Reading} consists of $32\times32$ colored digit images. There are 73257 images in the training set, 26032 images in the test set and 531131 images in the extra set. We followed the common practice \cite{He2016Deep, Huang2016Deep, Huang2017Densely, Lee2014Deeply} that the training set and the extra set were combined to train the network.

{\bf ImageNet:} The ILSVRC 2012 dataset \cite{Deng2009ImageNet, russakovsky2015imagenet} contains around 1.2 million images for training, and 50,000 images for validation. Those images cover 1000 different classes. We report the classification errors on the validation set.

\begin{table*}[htb]
\centering
\caption{Test Error Rates of Various Models on CIFAR and SVHN (\%).}
\label{tab:compare-cifar10-svhn}
\begin{tabular}{cc ccccc}
\toprule
Model &Depth & Params & CIFAR10& CIFAR100& SVHN  \\
\toprule
All-CNN \cite{springenberg2014striving} &- & - & 7.25& 33.71& -  \\
Deeply Supervised Net \cite{Lee2014Deeply}  &- & - & 7.97& 34.57& 1.92 \\
Highway Network \cite{Srivastava2015Training}  &19 & 2.3 M & 7.72& 32.39& -  \\
FractalNet \cite{Larsson2017FractalNet} &21& 38.6M & 5.22& 23.30& 2.01 \\
FractalNet+Dropout+Drop-path \cite{Larsson2017FractalNet}  &21& 38.6M  & 4.60& 23.73& 1.87 \\
ResNet with Stochastic Depth \cite{Huang2016Deep} &110& 1.7M  &5.23& 24.58& 1.75 \\
ResNet with Stochastic Depth \cite{Huang2016Deep} &1202& 10.2M &4.91& -& -\\
Wide ResNet \cite{zagoruyko2016wide}  &16& 11.0M  &4.81& 22.07& -\\
Wide ResNet with Dropout \cite{zagoruyko2016wide}  &16& 2.7M  &-& -&  1.64  \\
ResNet (pre-activation) \cite{He2016Identity}  &164& 1.7M  &5.46& 24.33& -\\
ResNet (pre-activation) \cite{He2016Identity} &1001& 10.2M  &4.62& 22.71& -\\
DenseNet   (k = 12) \cite{Huang2017Densely}        &100 & 7.0M  & 4.10& 20.20& 1.67 \\
DenseNet-BC (k = 12) \cite{Huang2017Densely}        &100 & 0.8M  &4.51& 22.27& - \\
DenseNet   (k = 24) \cite{Huang2017Densely}        &100 & 27.2M  & 3.74& 19.25& 1.67 \\
ResNet \cite{He2016Deep}  &110& 1.7M & 6.61& -& -  \\
ResNet (reported in \cite{Huang2016Deep})  &110& 1.7M  &6.41& 27.22& 2.01 \\
SE-ResNet-110 \cite{Hu2017Squeeze} & 110 & 1.9M &5.21 & 23.85 & - \\
SE-ResNet-110 \cite{Hu2017Squeeze} & 110 & 10.2M & 4.44  & 22.48  & 1.78 \\
SENet-29\cite{Hu2017Squeeze}& 29 & 35.0M &3.68 &17.78& - \\
ResNeXt-29, 16$\times$32d \cite{xie2016aggregated} & 29 & 25.2M & 3.87 & 18.56 & - \\
ResNeXt-29, 8$\times$64d \cite{xie2016aggregated} &29& 34.4M &3.65 &17.77& - \\
SKNet-29 \cite{Xiang2019SKNet} &29& 27.7M & 3.47 & 17.33 & - \\
SKNet-110* \cite{Xiang2019SKNet} &110& 25.3M & 3.56$\pm$0.08 & 18.34$\pm$0.06 & 1.56$\pm$0.03 \\
\hline
GRCNN-56     &56 & 5.8M   & 4.01$\pm$0.11 & 19.97$\pm$0.08 & 1.63$\pm$0.05 \\
GRCNN-110    &110& 10.6M  & 3.82$\pm$0.10 & 19.45$\pm$0.07   & 1.61$\pm$0.04 \\
SK-GRCNN-110  &110 & 25.7M  & {\bf3.42$\pm$0.04} & 18.18$\pm$0.09 &  {\bf1.55$\pm$0.02} \\
\bottomrule
\end{tabular}
\begin{flushleft}
  \quad{\quad\quad \quad\quad\quad\quad\quad\quad\quad \it * denotes our re-implementation.}
  \end{flushleft}
\end{table*}

\begin{table}[htb]
\centering
\caption{Inference time of different models on the test set of CIFAR-10 (seconds).}
\label{tab:cifar10-time}
\begin{tabular}{cc ccccc}
\toprule
Model &Depth & Params  & Time \\
\toprule

ResNet \cite{He2016Deep}  &110& 1.7M  & 10.87 \\
SE-ResNet-110 \cite{Hu2017Squeeze} & 110 & 1.9M & 11.47 \\
SE-ResNet-110 \cite{Hu2017Squeeze} & 110 & 10.2M & 17.97 \\
DenseNet   (k = 12) \cite{Huang2017Densely}        &100 & 7.0M  &  29.10 \\
DenseNet-BC (k = 12) \cite{Huang2017Densely}        &100 & 0.8M  & 24.38 \\
GRCNN-110       &109& 10.6M  & 22.82 \\
\bottomrule
\end{tabular}
\end{table}

\subsection{Implementation Details on CIFAR and SVHN}\label{sec:implement_small}

\begin{figure*}
\centering
\includegraphics[width=0.8\linewidth]{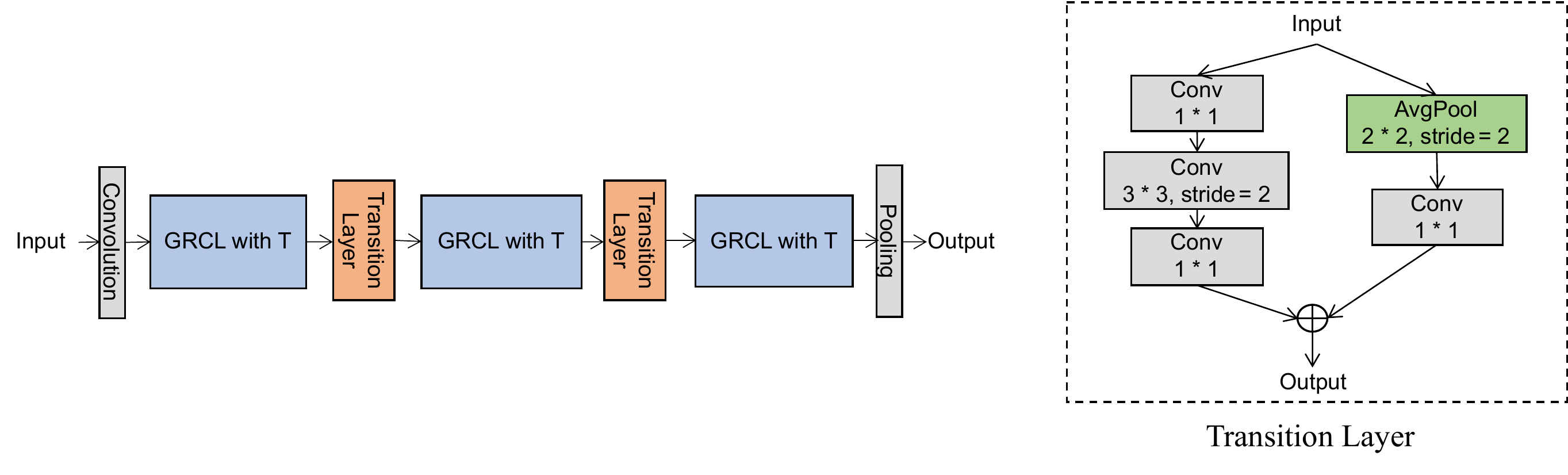}
\caption{Left: GRCNN with three GRCL blocks for object Recognition on CIFAR and SVHN. The adjacent blocks are connected by a transition layer that is responsible for downsampling. Right: The transition layer which is similar to the bottleneck layer used for downsampling in \cite{He2016Deep, He2016Identity}. For clarity, batch normalization layers and activation layers are omitted.}
\label{fig:smallnet}
\end{figure*}

The overall network architecture for the three small datasets, CIFAR-10, CIFAR-100 and SVHN, is shown in Fig.~\ref{fig:smallnet}. The input image with size $32\times32$ was fed into the first convolutional layer which consisted of 64 convolutional kernels with size $3\times3$ and stride 1. It followed three GRCLs interleaved with two ``transition layers''. As Fig.~\ref{fig:smallnet} shows, each transition layer is also implemented by a 3-layer bottleneck block, which is similar to the bottleneck blocks used for downsampling in \cite{He2016Deep, He2016Identity}.
The last layer preceding the output layer was a global average pooling layer. For RCNN, the GRCL blocks were replaced with the corresponding RCL blocks.

We trained the models using the conventional stochastic gradient descent (SGD) with batch size 64. The training epochs were 40 and 300 on SVHN and CIFAR, respectively. The learning rate was 0.1 initially, and it was multiplied by 0.1 at 50\% and 75\% of the total training epochs.  For preprocessing, we normalized the input data by using channel mean and standard deviation. On the CIFAR datasets, we evaluated the models under the data augmentation protocol. The augmentation techniques were mirroring and translation, which have been broadly used \cite{He2016Deep, Huang2016Deep, Larsson2017FractalNet, Huang2017Densely, Lee2014Deeply}. As for SVHN, no data augmentation was used. To avoid over-fitting, following previous works \cite{Huang2016Deep, zagoruyko2016wide} we inserted a dropout layer in GRCNN. Similar to \cite{Huang2017Densely}, the models saved at the end of training were evaluated. Five independent runs were performed for every model we investigated, and the mean error rates will be reported.

\begin{table*}[ht]
\centering
\caption{GRCNN Configuration for ImageNet Classification.}
\label{tab:arch-imagenet}
\begin{tabular}{cccccc}
\toprule
Layer & Output size & GRCNN-55 & GRCNN-109& SK-GRCNN-55 & SK-GRCNN-109\\
\toprule
Convolution & $112\times112$ &  \multicolumn{4}{c}{$7\times7$ conv, stride 2, feature maps: 64} \\
\hline
Pooling & $56\times56$ & \multicolumn{4}{c}{$3\times3$ max pooling, stride 2} \\
\hline
Convolution & $56\times56$ &  \multicolumn{4}{c}{$3\times3$ conv, stride 1, feature maps: 64} \\
\hline
\multirow{4}{*}{GRCL \#1}& \multirow{4}{*}{$56\times56$}&Iterations: 3&Iterations: 3 & Iterations: 3  & Iterations: 3\\
      &     & \Bigg[ \tabincell{c}{$1\times1$, 64 \\ $3\times3$, 64 \\ $1\times1$, 256} \Bigg]& \Bigg[ \tabincell{c}{$1\times1$, 64 \\ $3\times3$, 64 \\ $1\times1$, 256} \Bigg] &
      \Bigg[ \tabincell{c}{$1\times1$, 128 \\ SK[M=2, G=32, r=16], 128 \\ $1\times1$, 256} \Bigg]
      & \Bigg[ \tabincell{c}{$1\times1$, 128 \\ SK[M=2, G=32, r=16], 128 \\ $1\times1$, 256} \Bigg]\\
\hline
\multirow{2}{*}{Transition \#1}& \multirow{2}{*}{$28\times28$} & \multicolumn{4}{c}{\multirow{2}{*}{Transition Layer, output feature maps: 256}}  \\
                               &       &       \\
\hline
\multirow{4}{*}{GRCL \#2}& \multirow{4}{*}{$28\times28$}&Iterations: 3& Iterations: 3 & Iterations: 3 &  Iterations: 3 \\
      &     &\Bigg[ \tabincell{c}{$1\times1$, 128 \\ $3\times3$, 128 \\ $1\times1$, 512} \Bigg]&\Bigg[ \tabincell{c}{$1\times1$, 128 \\ $3\times3$, 128 \\ $1\times1$, 512} \Bigg] & \Bigg[ \tabincell{c}{$1\times1$, 256 \\ SK[M=2, G=32, r=16], 256 \\ $1\times1$, 512} \Bigg]  &\Bigg[ \tabincell{c}{$1\times1$, 256 \\ SK[M=2, G=32, r=16], 256 \\ $1\times1$, 512} \Bigg] \\
\hline
\multirow{2}{*}{Transition \#2}& \multirow{2}{*}{$14\times14$} & \multicolumn{4}{c}{\multirow{2}{*}{Transition Layer, output feature maps: 512}}  \\
                               &       &       \\
\hline
\multirow{4}{*}{GRCL \#3}& \multirow{4}{*}{$14\times14$}&Iterations: 4&Iterations: 22 & Iterations: 4 & Iterations: 22 \\
      &     &\Bigg[ \tabincell{c}{$1\times1$, 256 \\ $3\times3$, 256 \\ $1\times1$, 1024} \Bigg]& \Bigg[ \tabincell{c}{$1\times1$, 256 \\ $3\times3$, 256 \\ $1\times1$, 1024} \Bigg]& \Bigg[ \tabincell{c}{$1\times1$, 512 \\ SK[M=2, G=32, r=16], 512 \\ $1\times1$, 1024} \Bigg]&\Bigg[ \tabincell{c}{$1\times1$, 512 \\SK[M=2, G=32, r=16], 512 \\ $1\times1$, 1024} \Bigg]\\
\hline
\multirow{2}{*}{Transition \#3}& \multirow{2}{*}{$7\times7$} & \multicolumn{4}{c}{\multirow{2}{*}{Transition Layer, output feature maps: 1024}}  \\
                               &       &       \\
\hline
\multirow{4}{*}{GRCL \#4}& \multirow{4}{*}{$7\times7$} &Iterations: 3&Iterations:  3 & Iterations: 3 & Iterations: 3\\
      &     &\Bigg[ \tabincell{c}{$1\times1$, 512 \\ $3\times3$, 512 \\ $1\times1$, 2048} \Bigg] & \Bigg[ \tabincell{c}{$1\times1$, 512 \\ $3\times3$, 512 \\ $1\times1$, 2048} \Bigg]&\Bigg[ \tabincell{c}{$1\times1$, 1024 \\  SK[M=2, G=32, r=16], 1024 \\ $1\times1$, 2048} \Bigg] & \Bigg[ \tabincell{c}{$1\times1$, 1024 \\ SK[M=2, G=32, r=16], 1024 \\ $1\times1$, 2048} \Bigg]\\
\hline
\multirow{2}{*}{Classification} & $1\times1$ &\multicolumn{4}{c}{$7\times7$ global average pooling}\\
     &   &\multicolumn{4}{c}{1000 dimensional fully-connected, softmax}\\
\bottomrule
\end{tabular}
\begin{flushleft}
\quad\quad {\it Note that, as for each transition layer in SK-GRCNNs, the $3\times3$ convolutional layer is replaced by a SKConv \cite{Xiang2019SKNet} with stride 2 for fair comparison.}
\end{flushleft}
\end{table*}
\subsection{Model Analyses on CIFAR-10}\label{sec:model-analyses}
To investigate the effects of components or hyper-parameters of the proposed model on the classification performance on the CIFAR-10 dataset, a couple of architectures were designed differing in the number of iterations $T$, with and without gates, and so on (See Table~\ref{tab:compare-cifar10}). The recurrent transformation $\mathcal{T}^\text{R}$ of GRCL in each iteration is implemented by a simple ``BN-ReLU-Conv''.
Therefore,
in Table~\ref{tab:compare-cifar10}, $(a,b,c)$ in the first column indicates that the three GRCLs in Fig. \ref{fig:smallnet} had $a, b$ and $c$ iterations, respectively. $(a,b,c)$ in the second column indicates that the three GRCLs in Fig. \ref{fig:smallnet} had $a, b$ and $c$ feature maps, respectively. The group number of grouped convolution layers used in the improved GRCNN is set to 16. For simplicity, here the same $T$ was used in all three GRCLs or RCLs. Table \ref{tab:compare-cifar10} reports the mean and standard deviation of the test error rates in percentage over five independent runs for each architecture.

{\bf Effect of the gates.} We first compared the original GRCNN described in Section \ref{sec:originalGRCNN} with the RCNN described in Section \ref{sec:RCNN}. Both models used tied weights, {\it i.e.}, $w^\text{R}(t)$ and $w^\text{R}_\text{g}(t)$ were identical across $t$. The only difference is that the former had gates on the recurrent connections while the latter did not. When the number of iterations $T$ in each original GRCL increased from 2 to 5, the test error rate of the original GRCNN decreased gradually (Table \ref{tab:compare-cifar10}). However, this trend was only observed in the RCNN when $T\le3$. When $T>3$, the test error rate of the RCNN increased. These results indicated that increasing RF may not always help the RCNN, and sometimes it may even hurt the performance of the RCNN. Due to the adaptive control of RF in GRCL, increasing $T$ always helped the original GRCNN.

\begin{figure}
\includegraphics[width=\linewidth]{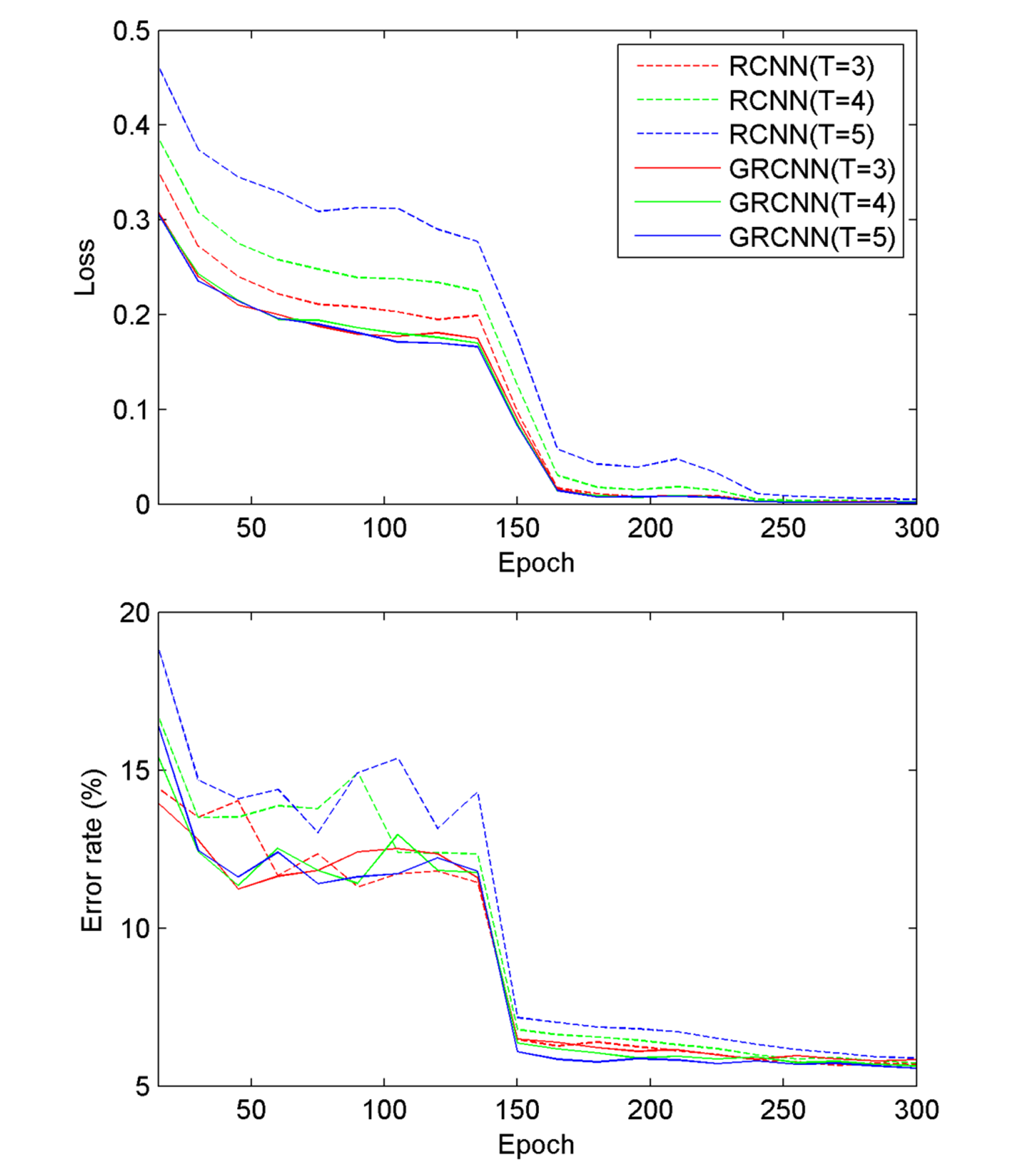}
\caption{Loss curves (upper) and test error rate curves (lower) of the RCNN with weight sharing and the original GRCNN with weight sharing. The horizontal axis represents the training epoch. }
\label{fig:training-cifar10}
\end{figure}

The training loss curves and test error rate curves of the RCNN and the original GRCNN with $T = 3, 4, 5$ are plotted in Fig.~\ref{fig:training-cifar10}.  With larger $T$, RCNN converged slower and the final loss became higher. However, the original  GRCNN with different iterations had similar training curves and the converged losses were all lower than those of RCNN. The results indicated that the original  GRCNN had better training performance than the RCNN in terms of speed and accuracy. The test error rate curves indicated that, with the same $T$, the original GRCNN always achieved lower test error than RCNN.

{\bf Effect of the accumulated contributions over iterations.}
The original GRCL \eqref{eqn:GRCL-nips} uses the gated recurrent contribution in the last iteration while the improved GRCL (equation \eqref{eqn:GRCL}) uses the gated recurrent contributions in all iterations.
With the same number of iterations in each GRCL, the improved GRCNN achieved lower accuracies than the original GRCNN  (Table~\ref{tab:compare-cifar10}), which indicated the advantage of the accumulated contributions over all iterations.

{\bf Effect of the number of recurrent iterations.}
When the number of iterations increased from 2 to 5, the mean test error rate of the original GRCNN decreased gradually from 5.92\% to 5.57\%, and the mean test error rate of the improved GRCNN decreased gradually from 5.83\% to 5.49\%. The results suggested that more iterations are more advantageous for both models.

{\bf Effect of weight sharing.}
We then trained and tested the RCNN, the original GRCNN and the improved GRCNN with untied weights, {\it i.e.}, $w^\text{R}(t)$ and $w^\text{R}_\text{g}(t)$ were different for different $t$. The conclusions about the effects of the gates, the accumulated contributions over iterations and the number of iterations drew on the recurrent networks were also valid (Table~\ref{tab:compare-cifar10}). In addition, the architectures with untied weights always achieved lower test errors than their counterparts with tied weights. This is reasonable because without weight sharing, the model has much more parameters and therefore higher capability to model the mapping from images to labels.

\begin{figure}
\centerline{\includegraphics[width=\linewidth]{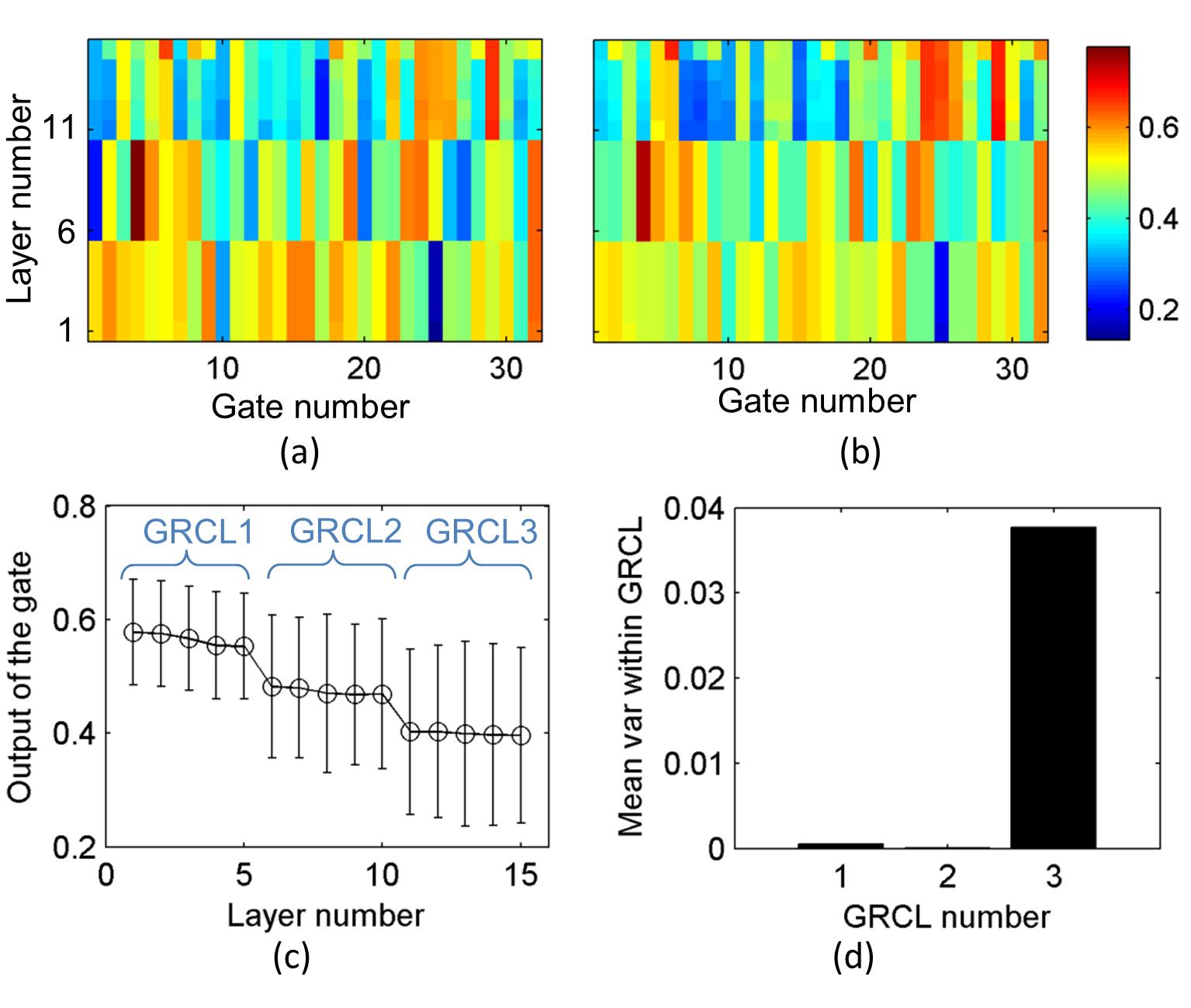}}
\caption{Outputs of the gates in GRCNN given CIFAR-10 images. (a,b) The output values of the gates in different layers given two sample input images. Every layer had 128 gates  but for clarity only the first 32 gates in every layer are shown. The 15 layers belong to three GRCLs (five per GRCL). Other layers in the model including intermediate layers between GRCLs are not shown. Note that the gates with the same index over 128 in different GRCLs do not have a direct relation. That is why the strips break at the boundaries of different GRCLs.  (c) The mean output of the gates over 128 gates and 10,000 training images. Error bars indicate the standard deviation. (d) The mean of the variance of the gate outputs within each GRCL.}
 \label{fig:gate-output}
\end{figure}

{\bf Behavior of the gates.}
Given an input image, how do these gates collaborate with each other to give an output? We then analyzed the output values of the gates in the GRCNN in the last row of Table \ref{tab:compare-cifar10}. Since there were three GRCLs and each one had five iterations, the model had 15 gated layers. Note that every feature map had one gate, then we had $128\times15$ gates in total. Fig. \ref{fig:gate-output}a,b show the output values of some gates given two sample input images. It is seen that the gates in GRCL1 tended to output larger values while the gates in GRCL3 tended to output smaller values. Statistical results over 10,000 input images verified this point, as shown in Fig. \ref{fig:gate-output}c. This makes sense as neurons in lower layers had small RFs and they tended to incorporate more context for better understanding of the context, while neurons in higher layers already had large RFs and they tended to focus on local regions that help the model to recognize specific objects. In addition, Fig. \ref{fig:gate-output}c shows that the variances of the outputs increased from GRCL1 to GRCL3, which indicates that the gates were more and more selective to inputs.

The stripe pattern in Fig. \ref{fig:gate-output}a,b indicates that the five gates in the same GRCL, especially GRCL1 and GRCL2, at the same location (same index over 128 gates in a layer) had similar outputs. But this pattern was less prominent in GRCL3. To provide a quantitative measure, we calculated the variance of the outputs of the five gates with the same index in each GRCL given an input image, then averaged over all 128 gates and 10,000 input images. The average in GRCL3 was much larger than the average in GRCL1 and GRCL2 (Fig. \ref{fig:gate-output}d). The results
indicate that the gates in GRCL1 and GRCL2 in different iterations had similar modulation effects to their inputs while the gates in GRCL3 had different modulation effects to their inputs.

To sum up, the observations in these experiments are summarized as follows.
\begin{itemize}
\item The improved GRCNN performed better than the original GRCNN and the RCNN.
\item More iterations in the improved GRCL led to better results.
\item The improved GRCNN performed better without weight sharing than with weight sharing.
\end{itemize}
In order to compete with existing models in terms of accuracy, we need to build deep GRCNNs which have many iterations. To save the number parameters, for each GRCL, we only untie the parameters in the recurrent path, say, $w^R$, but share the parameters of gates, say, $w_\text{g}^R$ in \eqref{eqn:gate}, among different iterations. Hereafter, without specification, GRCNN refers to this particular model.

\begin{table}
\centering
\caption{Top-1 and Top-5 Error Rates on the ImageNet-2012 Validation with Different Popular Architectures (Single-Crop Test Error Rates) (\%). }
\label{tab:imgnet}
\begin{tabular}{ccccc}
\toprule
Model & {\footnotesize GFLOPS} & Param &Top-1 & Top-5 \\
\toprule
ResNet-50 \cite{He2016Deep}& 3.80   &25.6M &24.70  & 7.80 \\
ResNet-101 \cite{He2016Deep}& 7.60 &44.6M &23.60  & 7.10 \\
ResNet-152 \cite{He2016Deep}& 11.3 &60.2M &23.00  & 6.70 \\
DenseNet-121 \cite{Huang2017Densely} & 2.90 &8.4M &25.02  & 7.71 \\
DenseNet-169 \cite{Huang2017Densely} & 3.44 &14.2M  &23.80  & 6.85 \\
DenseNet-201 \cite{Huang2017Densely} & 4.39 &20.0M &22.58 & 6.34 \\
SE-ResNet-50 \cite{Hu2017Squeeze} & 4.25 &27.7M &23.29  & 6.62 \\
SE-ResNet-101 \cite{Hu2017Squeeze} & 8.00 &49.2M &22.38  & 6.07\\
SKNet-50 \cite{Xiang2019SKNet}& 4.47 &27.5M & 20.79  & - \\
SKNet-100 \cite{Xiang2019SKNet} & 8.46 &48.9M  & 20.19 & - \\
\hline
RCNN-24 (our implem.) & 2.51 &25.6M &29.43  & 10.08  \\
RCNN-55 (our implem.) & 3.36 &25.8M &32.92  & 13.64  \\
GRCNN-55 & 3.73 &24.9M  & 23.12   &  6.52  \\
GRCNN-109 & 7.80 &45.1M & 21.95   &  5.95  \\
SKNet-50 (our implem.)& 4.47 &27.5M & 20.88  & 5.11 \\
SKNet-100 (our implem.)& 8.46 &48.9M  & 20.27 & 4.89 \\
SK-GRCNN-55 & 4.39 &27.4M  & 20.68  &  5.02  \\
SK-GRCNN-109 & 8.45 &50.0M & {\bf20.09}   &  {\bf 4.86}  \\
\bottomrule
\end{tabular}\\
\vspace{2mm}
\end{table}

\begin{table}
\centering
\caption{Compared with Two Models Having Adaptive Receptive Fields in Terms of Error Rate
on the ImageNet-2012 Validation Set with Single-Crop Protocol (\%). }
\label{tab:deform}
\begin{tabular}{ccc}
\toprule
Model & Top-1 & Top-5 \\
\toprule
Saliency Network ResNet-50 \cite{recasens2018learning} & 23.52 &  6.92  \\
Deformable ResNet-50 \cite{dai2017deformable} &   23.40  & 6.80 \\
GRCNN-55 & 23.12  & 6.52  \\
Deformable GRCNN-55 & 22.91  & 6.48  \\
Deformable v2 GRCNN-55 & {\bf21.63}  & {\bf6.19}  \\
\hline
Saliency network ResNet-101 \cite{recasens2018learning} &  22.32 & 6.19 \\
Deformable ResNet-101 \cite{dai2017deformable} &  21.60   & 5.80  \\
GRCNN-109 & 21.95 & 5.95  \\
Deformable GRCNN-109  &  21.35   & 5.74  \\
Deformable v2 GRCNN-109 & {\bf20.62}  & {\bf5.38}  \\
\bottomrule
\end{tabular}\\
\vspace{2mm}
\begin{flushleft}
\end{flushleft}
\end{table}

\subsection{Comparison with Existing Methods on Small Datasets}\label{sec:cifar-sota}
In order to compare GRCNN with existing models, we experimented with three deeper architectures as illustrated in Fig. \ref{fig:smallnet}:
\begin{itemize}
 \item GRCNN-56: The number of iterations (3, 5, 7), the number of feature maps (128, 160, 192).
 \item GRCNN-110: The number of iterations (6, 9, 18), the number of feature maps (128, 160, 192).
 \item SK-GRCNN-110: The number of iterations (6, 9, 18), the number of feature maps (192, 320, 448).
\end{itemize}

The number in the name of each architecture denotes the depth. Unlike the settings in Sections \ref{sec:model-analyses}, the $\mathcal{T}^\text{R}$ of GRCL in each iteration was implemented by a general 3-layer bottleneck block \cite{He2016Deep, He2016Identity} or a 3-layer bottleneck block with SKConv \cite{Xiang2019SKNet}, in order to build deep models.
The depth of each architecture can be calculated accordingly. For example, with $T=3, 5$ and $7$, the three GRCLs in Fig. \ref{fig:smallnet} had $1+3\times 3$, $1+5\times 3$ and $1+7\times 3$ layers, respectively. There were two 3-layer transition layers and one convolutional layer between the input and GRCLs as depicted in Fig. \ref{fig:smallnet}, plus one output layer. Therefore, the depth of the first architecture was 56. The depth of the other two architectures can be calculated in the same way.
For each architecture, the three numbers of feature maps listed above denote the number of $3\times3$ convolutional filters in bottleneck blocks of each GRCL respectively. We defined the ``expansion rate'' of a 3-layer bottleneck block as the ratio of the number of the third $1\times1$ convolutional filters to the number of the first $1\times1$ convolutional filters. The expansion rate of the bottleneck block was set to 4 and 2 for GRCNN and SK-GRCNN, respectively. The group number of $w^\text{F}$, $w^\text{F}_\text{g}$ and $w^\text{R}_{\text{g}}(t)$ in GRCNN and SK-GRCNN were all set to 16.
The hyper-parameters of SKConv in SK-GRCNN-110 were exactly the same
as those used in the CIFAR experiment in \cite{Xiang2019SKNet}.

The results of the three architectures on CIFAR-10, CIFAR-100 and SVHN are presented in Table \ref{tab:compare-cifar10-svhn}.
The mean and standard deviation of the test error rates over five runs are reported. GRCNN-110 outperformed GRCNN-56 on CIFAR and SVHN and SK-GRCNN-110 outperformed GRCNN-110. In addition, SK-GRCNN-110 outperformed SKNet-110 on the CIFAR datasets, indicates the usefulness of the proposed gating mechanism. Finally, the SK-GRCNN achieved competitive results compared with the state-of-the-art models.

Table~\ref{tab:cifar10-time} presents the inference time of GRCNN-110 and several popular networks with similar depth listed in Table~\ref{tab:compare-cifar10-svhn} on the CIFAR-10 test set (tested on a single GeForce GTX TITAN X GPU with PyTorch \cite{NEURIPS2019_9015}). The models were run with batch size 64 and the total time on the test set was reported. From Tables~\ref{tab:compare-cifar10-svhn} and \ref{tab:cifar10-time}, it is seen that though the GRCNN and DenseNet can achieve lower error rates, they are not as efficient as the ResNet and its variants.  This is a shortage of both GRCNN and DenseNet.

\begin{table*}
  \caption{The Configuration of GRCNN in Scene Text Recognition.}
  \label{Network Settings-ocr-large}
  \centering
  \begin{tabular}{ccccccccccc}
    \toprule
Conv      & GRCL & Conv  &GRCL  & Conv    &GRCL       & Conv  &   GRCL     & Conv &  GRCL & Conv  \\
\toprule
$3\times3$& $3\times3$  &$2\times2$  &$3\times3$ & $2\times2$ & $3\times3$&$2\times2$& $3\times3$ & $2\times2$& $3\times3$& $2\times2$\\
  num: 32 & num: 32 & num: 32      &num:  64 &  num:  64    &  num: 128 &  num: 128   &  num: 256  &  num: 256    &  num: 512 &num: 512\\
sh:1 sw:1 & sh:1 sw:1 &sh:2 sw:2      &  sh:1 sw:1 &    sh:2 sw:2 & sh:1 sw:1 &   sh:2 sw:1  & sh:1 sw:1  & sh:2 sw:1 & sh:1 sw:1&sh:2 sw:1 \\
ph:1 pw:1 & ph:1 pw:1 & ph:0 pw:0    &ph:1 pw:1 &    ph:0 pw:0  & ph:1 pw:1 &   ph:0 pw:1  & ph:1 pw:1&   ph:0 pw:1& ph:1 pw:1&ph:0 pw:1 \\
    \bottomrule
  \end{tabular}
  \begin{flushleft}
  sh and sw: {\it the strides of the kernel along the height and width respectively}; ph and pw: {\it the padding values of height and width respectively}; num: {\it the number of convolutional filters}.
  \end{flushleft}
\end{table*}

\subsection{Implementation Details on ImageNet}\label{sec:implem_imagenet}
To verify the generalization of the proposed gating mechanism, we designed GRCNN based frameworks with 4 GRCLs, including GRCNNs and SK-GRCNNs. The input image size was $224\times224$. The first layer was a convolutional layer that had 64 convolutional kernels of size $7\times7$ with stride 2 and zero-padding 3. It was followed by a max pooling layer with size $3\times 3$ and stride 2, and another convolutional layer that had 64 convolutional kernels of size $3\times3$ with stride 1 and zero-padding 1. Then four GRCLs were used in sequence. Each of the first three GRCLs was followed by a transition layer.
In GRCNNs, $\mathcal{T}^\text{R}$ in each iteration and the transition layer are implemented by the general 3-layer bottleneck block \cite{He2016Deep, He2016Identity} (See the descriptions in sec.\ref{sec:implement_small}). As for SK-GRCNNs, $\mathcal{T}^\text{R}$ and the transition layer are implemented by the 3-layer bottleneck block with SK convolution \cite{Xiang2019SKNet}. The group number of grouped convolution layers used in the GRCNNs and SK-GRCNNs are all set to 32.
See Table~\ref{tab:arch-imagenet} for details.

For comparison, we also evaluated two versions of RCNN with 55 and 24 layers, respectively.  The RCNN-55 was obtained by replacing GRCLs in the GRCNN-55 with RCLs using general 3-layer bottleneck blocks. The RCNN-24 had a similar architecture, but the iteration number of every RCL was 2, and only one  ``BN-ReLU-Conv'' layer with $3\times3$ convolutional filters was used to implement $\mathcal{T}^\text{R}$. The number of output feature maps for the four RCLs were 128, 256, 512 and 1024, respectively.

During training, the input size is $224\times224$ and we use the same data augmentation method as \cite{Hu2017Squeeze}. The batch size was set to 256 and the models were trained for 100 epochs. Top-1 and Top-5 error rates were reported. The learning rate was set to 0.1 initially, and it was multiplied by 0.1 after every 30 training epochs. 
Note that we cannot reproduce the results of SKNets reported in the original paper \cite{Xiang2019SKNet}. To achieve the comparable results in the original paper, our reproduced SKNets were trained for 120 epochs by using
cosine learning rate decay and Mixup \cite{Hongyi2018}, and SK-GRCNNs also follow this setting. According to previous works \cite{Sutskever2013On}, the weight decay and Nesterov momentum were set to 10$^{-4}$ and 0.9, respectively. The label smoothing strategy \cite{Szegedy2015Rethinking} was used, and the network was initialized using the method in \cite{He2015Delving}.

During testing, we performed single-crop testing, and the input size is $224\times224$. To be specific, for the single-crop protocol, the center crop of a test image was used as the input.

\subsection{Results on ImageNet}\label{sec:results_imagenet}

The test error rates of the GRCNNs and RCNNs are presented in Table \ref{tab:imgnet}. First, RCNN-55 performed worse than a shallower mode RCNN-24. This observation is consistent with the results on CIFAR-10 (See Table \ref{tab:compare-cifar10}). Second, the three GRCNNs performed much better than RCNN-55.

Table \ref{tab:imgnet} also lists the results of some popular models including ResNet, DenseNet, SE-ResNet and SK-Net. GRCNN outperformed other models which also used the general 3-layer bottleneck block, i.e., ResNet and SE-ResNet. Note that the performance gap between ResNet and GRCNN is remarkable, which verifies the effectiveness of the proposed gating method since GRCNN is similar to ResNet equipped with the proposed gates.
When the SK convolution was introduced, SK-GRCNN achieved a little bit higher accuracy than SK-Net as well. 
Note that some models including ResNet with pre-activation \cite{He2016Identity}, Inception-V3 \cite{Szegedy2015Rethinking}, Inception-ResNet-v1 \cite{Szegedy2016Inception}, and Inception-V4 \cite{Szegedy2016Inception} used different test settings, e.g., larger input crops than $224\times224$ and 144-crop testing, and therefore are not compared in the table.

Finally, we compared the GRCNN with two models, the Deformable CNN \cite{dai2017deformable} and the Saliency Network \cite{recasens2018learning}, which also have adaptive RFs, and Table~\ref{tab:deform} presents the results. The GRCNN-55 and the GRCNN-109 obtained slightly better results than the Saliency Network ResNet-50 and the Saliency Network ResNet-101, respectively.  The GRCNN-55 performed better than the Deformable ResNets-50 while the GRCNN-109 performed worse than the Deformable ResNets-101. The performance of GRCNN was boosted by using the deformable convolutional layers (Table~\ref{tab:deform}). We built a deformable GRCNN by introducing deformable convolution layers into the bottleneck of stage GRCL\#4 in GRCNN. As for deformable v2 GRCNN, we followed \cite{zhu2019deformable} by applying deformable convolution layers at the bottleneck of stages GRCL\#2$\sim$GRCL\#4 in GRCNN. The improved performance indicates that our method and the deformable convolution are complimentary for implementing adaptive RFs to improve the performance. This phenomenon was also observed in \cite{zhu2019deformable}, which introduces sigmoid layers to obtain the modulation scalars for achieving effective RFs.

\section{Scene Text Recognition}\label{sec:OCR}

\begin{figure}
\includegraphics[width=\linewidth]{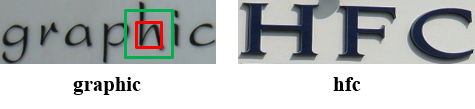}
\caption{Two sample images from the ICDAR2003 dataset. In the first example, for recognizing the character ``h'', the classical RF (red square) and desired non-classical RF (green) of GRCNN are shown. If the non-classical RF covers unrelated information, such as the characters
``p'' and ``i'', it will distract the model.}
\label{fig:IC_example}
\end{figure}

The second application of the proposed GRCNN is OCR. In this application, there are multiple characters in one image to be recognized (Fig. \ref{fig:IC_example}). Sometimes, the gap between characters is large and sometimes the gap is small. It is desired that a model can automatically adjust its focus to the characters.
We expect that GRCNN suits this application because its built-in gates enable it to do so.

Unlike object recognition, for which a backbone CNN alone suffices to obtain good results, OCR requires additional techniques to process the character sequences. Many sophisticated pipelines have been proposed \cite{Shi2016An, shi2018aster}, where CNN backbones play an important role. Since our aim was to examine the usefulness of the GRCNN in this application, we replaced the original backbone CNNs with the GRCNN in these frameworks and compared the results.

\begin{table*}
\caption{Scene Text Recognition Accuracies of Different Models.}
\label{tab:ocr}
\centering
\begin{tabular}{cccccccccc}
    \toprule
    Method  & Data  & SVT-50  & SVT  & IIIT5K-50  & IIIT5K-1k   & IIIT5K   & IC03-50   & IC03-Full  & IC03  \\
    \toprule
    ABBYY \cite{Wang2012End} & - & 35.0\%  & - & 24.3\%  & -   & -  & 56.0\%   & 55.0\%  & -  \\
    wang et al. \cite{Wang2012End}& - & 57.0\%  & - & -  & -   & -  & 76.0\%   & 62.0\%  & -  \\
    Mishra et al. \cite{Mishra2013Scene}& - & 73.2\%  & - & -  & -   & -  & 81.8\%   & 67.8\%  & -  \\
    Novikova et al. \cite{Novikova2012Large}&-  & 72.9\%  & - & 64.1\%  & 57.5\%  & -  & 82.8\%   & -  & -  \\
    wang et al. \cite{Wang2013End}&- & 70.0\%  & - & -  & -  & -  & 90.0\%   & 84.0\%  & -  \\
    Bissacco et al. \cite{Bissacco2014PhotoOCR}&- & 90.4\%  & 78.0\%& -  & -  & -  & -  & - & -  \\
    Goel et al. \cite{Goel2013Whole}&- & 77.3\%  & - & -  & -  & -  & 89.7\%  & - & -  \\
    Alsharif \cite{Alsharif2013End}&- & 74.3\%  & - & -  & -  & -  & 93.1\%  &88.6\%  & -  \\
    Almazan et al. \cite{Almaz2014Word}&- & 89.2\%  & - & 91.2\% & 82.1\%  & -  & -  &-  & -  \\
    Lee et al. \cite{Lee2014Region}&- & 80.0\%  & - & - & -  & -  & 88.0\%  & 76.0\%  & -  \\
    Yao et al. \cite{Yao2014Strokelets}&- & 75.9\%  & - & 80.2\% & 69.3\%  & -  & 88.5\%  & 80.3\%  & -  \\
    Rodriguez et al. \cite{Rodriguez2015Label}&- & 70.0\%  & - & 76.1\% & 57.4\%  & -  & - & - & -  \\
    Jaderberg et al. \cite{Jaderberg2014Deep}&90k & 86.1\%  & - & - & -  & -  & 96.2\% & 91.5\% & -  \\
    Su and Lu et al. \cite{Su2014Accurate}&- & 83.0\%  & - & - & -  & -  & 92.0\% & 82.0\% & -  \\
    Gordo \cite{Gordo1998Supervised}& &90.7\% & - & 93.3\% & 86.6\%  & -  & - & - & -  \\
    Jaderberg et al. \cite{Jaderberg2014Deep1}&90k & 93.2\%  & 71.1\% & 95.5\% & 89.6\%  & -  & 97.8\% & 97.0\% & 89.6\%  \\
    Baoguang et al. \cite{Shi2016An}&90k & {\bf 96.4\%}  & 80.8\% & 97.6\% & 94.4\%  & 78.2\% & 98.7\% & 97.6\% & 89.4\%  \\
    Chen-Yu et al. \cite{Lee2016Recursive}&90k & 96.3\%  & 80.7\% & 96.8\% & 94.4\%  & 78.4\%  & 97.9\% & 97.0\% & 88.7\%  \\
     \hline
    ResNet-BLSTM  \cite{jianfeng2017deep}&90k & 96.0\%  &  80.2\% & 97.5\% & 94.9\%  & 79.2\%  & 98.1\% & 97.3\% & 89.9\%  \\
    Original GRCNN-BLSTM \cite{jianfeng2017deep}&90k & 96.3\%  &  81.5\% & 98.0\% & 95.6\%  & 80.8\%  & {\bf98.8\%} & {\bf97.8\%} & 91.2\%  \\
    GRCNN-BLSTM&90k & {\bf 96.4\%}  & {\bf 82.1\%} & {\bf 98.2\%} & {\bf 95.9\%}  & {\bf 81.8\%}  & {\bf 98.8\%} & {\bf 97.8\%} & {\bf 91.7\%}  \\
    \toprule
    FAN \cite{cheng2017focusing} &90k + ST & 97.1\% & 85.9\% & 99.3\% & 97.5\%  & 87.4\%  & {\bf99.2\%} & 97.3\% & 94.2\%  \\
    AON \cite{cheng2018aon}&90k + ST & 96.0\%  &  82.8\% & 99.6\% & 98.1\%  & 87.0\%  & 98.5\% & 97.1\% &  91.5\%  \\
    SAM \cite{liao2019mask}& 90k + ST & {\bf98.6\%} & {\bf90.6\%} & 99.4\% & 98.6\% & 93.9\% & 98.8\% & 98.0\% & 95.2\% \\
    ESIR \cite{zhan2019esir} & 90k + ST & 97.4\% & 90.2\% & {\bf 99.6\%} & {\bf98.8\%} & 93.3\% & - & - & - \\
    SE-ASTER \cite{qiao2020seed} & 90k + ST & - & 89.6\% & - & - & 93.8\% & - & - & - \\
    \hline
      ASTER with ResNet \cite{shi2018aster} &90k + ST &  97.4\% & 89.5\%& {\bf 99.6\%} & {\bf 98.8\%}  &  93.4\%  &  98.8\% & 98.0\%&  94.5\%  \\
      ASTER with GRCNN &90k + ST & 97.8\%  & {\bf 90.6\%} & 99.5\% & {\bf 98.8\%} & {\bf 94.0\%}  & 98.8\% & {\bf 98.2\%} & {\bf 95.6\%}  \\
     \bottomrule
  \end{tabular}\\
  \begin{flushleft}
  \quad{\it "50","1k" and "Full" appended to the dataset names
  denote the lexicon size used for the lexicon-based recognition task.
  The dataset name without a lexicon size indicates the unconstrained text
  recognition task.}
  \end{flushleft}
\end{table*}

\subsection{Datasets}
{\bf ICDAR2003:} ICDAR2003 \cite{Lucas2003ICDAR} contains 251 scene images and there are 860 cropped images of the words. We performed
unconstrained text recognition and constrained text recognition on this dataset.
Each image is associated with a 50-word lexicon defined by wang et al. \cite{Wang2012End}.
The full lexicon is composed of all per-image lexicons.

{\bf IIIT5K:} This dataset has 3000 cropped testing word images and 2000 cropped training
images collected from the Internet \cite{Shi2013Scene}. Each image has a lexicon of 50 words and a lexicon
of 1000 words.

{\bf Street View Text (SVT):} This dataset has 647 cropped word images from Google street view \cite{Wang2012End}. The 50-word lexicon defined by Wang et al \cite{Wang2012End} was used in our experiment.

{\bf Synth90k:} This dataset contains around 7 million training images, 800k validation
images and 900k test images \cite{Jaderberg2014Synthetic}. All of the word images were generated by a synthetic text
engine and are highly realistic.

{\bf SynthText:} This dataset was initially designed for text detection. However, it has been widely used for scene text recognition task by cropping text image patches according to ground-truth boxes. We followed \cite{shi2018aster} and cropped  7 million patches for training.

When evaluating the performance of our model on those benchmark
datasets, we followed the evaluation protocol in \cite{Wang2012End}. We performed recognition
on the words that contain only alphanumeric characters (A-Z and 0-9) and at
least three characters. All recognition results were case-insensitive.

\subsection{Pipelines}\label{sec:pipeline_OCR}
We conducted experiments on two prevalent scene text recognition frameworks. One is the Convolutional-Recurrent Neural Network (CRNN), and it consists of three parts: feature extraction, feature map splitting and sequence modeling. The feature extraction part can be performed by any CNN backbone to learn visual features from the input image. The second part slices the feature maps from left to right by column to form a feature sequence, which is then processed by the feature modeling part. The other one is ASTER framework \cite{shi2018aster} which is proposed to deal with the irregular scene text. It can be regarded as introducing a text rectification module before the CRNN and attentional LSTM decoders after the CRNN.
Both frameworks are supervised by Connectionist Temporal Classification (CTC) \cite{Graves2006Connectionist} loss during training. More details about these two frameworks can be found in their original papers.

\subsection{Implementation Details}\label{sec:implement_OCR}
The configuration of the GRCNN used in OCR is shown in Table~\ref{Network Settings-ocr-large}. Note that the backbones of CRNN, both ResNet and the original GRCNN, in \cite{jianfeng2017deep} had 20 parameterized layers while the ResNet backbone in ASTER \cite{shi2018aster} had 50 parameterized layers. Therefore, in our experiment, we built GRCNNs by keeping their depth comparable to those of backbones in these two pipelines for fairly comparison.

To construct a GRCNN-based CRNN, the feature extraction part was performed by the GRCNN and sequence modeling was performed by a Bidirectional LSTM with 256 units without dropout. The time steps for the GRCLs are set to 1, 3, 3, 3 and 3 separately, and only one ``BN-ReLU-Conv'' layer with $3\times3$ convolutional filters was used to implement $\mathcal{T}^\text{R}$ in each GRCL. The number of feature maps
is shown in Table~\ref{Network Settings-ocr-large}. The group number of grouped convolution layers used in GRCNN is set to 8.
The input was a gray-scale image with size 100$\times$32. Before inputting to the network, the pixel values were rescaled to the range (-1, 1). No further data augmentation methods were used. The ADADELTA method \cite{Zeiler2012ADADELTA} was used for training with the parameter $\rho$=0.9. The batch size was set to 128 and training stopped after 300k iterations. In order to fairly compare with our preliminary work \cite{jianfeng2017deep} and most other previous methods, all CRNN related models are only trained on Synth90k.

As for the ASTER pipeline \cite{shi2018aster}, the time steps of the GRCLs were 1, 2, 4, 4 and 3. To construct a deep GRCNN, we used the general 3-layer bottleneck block to implement $\mathcal{T}^\text{R}$ in each GRCL. Bottleneck blocks in each GRCL contained 32, 64, 128, 256, and 512 $3\times3$ convolutional filters separately, and the expansion rate of those bottleneck blocks was set to 2. The group number of grouped convolution layers used in GRCNN is set to 8.
The batch size was 64. The same ADADELTA method as described above was used for training, but a global learning rate was initially set to 1.0, and decayed to 0.1 and 0.01 at 0.6M and 0.8M iterations, respectively. This pipeline was trained on a combination of Synth90k and SynthText.

\subsection{Comparison with Existing Methods}
We evaluated GRCNN on several benchmark text recognition datasets in terms of the two pipelines mentioned in sec.\ref{sec:pipeline_OCR}, and we only pick the model which performs the best on the validation set of Synth90k for testing.

Table~\ref{tab:ocr} presents the results based on the CRNN pipeline. It is clear that the GRCNN-based CRNN outperformed most previous methods. In addition, with the comparable number of feature maps and depth, the GRCNN-based CRNN outperformed the ResNet-based CRNN. The GRCNN-based CRNN outperformed previous state-of-the-art methods which are only trained on Synth90k as well for both lexicon-based and lexicon-free based recognition.

Table~\ref{tab:ocr} also presents the results based on the ASTER pipeline. It is seen that the ResNet-based model performed worse than the GRCNN-based model. Moreover, the GRCNN-based ASTER achieved competitive results
when compared with previous state-of-the-art results.

\section{Object Detection}\label{sec:detection}


The third application of the proposed model is object detection. In this application, an image has multiple objects to be detected, and they may differ significantly in scale. See examples in Fig. \ref{fig:COCO_example}. The GRCNN may suit this application because its neurons have adaptive RFs.

\subsection{Datasets}
\begin{figure}
\includegraphics[width=\linewidth]{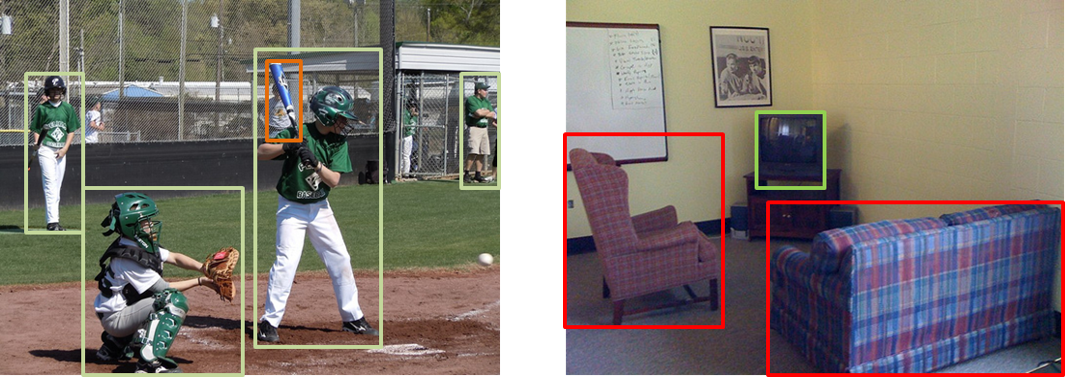}
\caption{Sample images from MS COCO dataset. The objects are cluttered together and differ significantly in scale. For example, in the first example, the baseball bat is very small compared with the player.}
\label{fig:COCO_example}
\end{figure}

The Microsoft common objects in context (MS COCO) \cite{lin2014microsoft} dataset contains 91 common object categories
and 82 categories have more than 5,000 labeled instances. The dataset consists of 2,500,000 labeled
instances in 328,000 images in total. Instances are segmented individually to aid in precise object localization. Several examples are shown in Fig. \ref{fig:COCO_example}. Following the evaluation protocol of MS COCO, we used the trainval35k set for training and evaluated the results on the \textit{test-dev} set.

\begin{table*}
\caption{Detection Results of Anchor-based One-stage Detectors on MS COCO \textit{Test-dev} Set.}
\label{tab:detect}
\centering
\begin{tabular}{ccc c cc c c c}
    \toprule[1pt]
    Method    & Backbone network  & AP  & AP$_{50}$   & AP$_{75}$  & AP$_{S}$   & AP$_{M}$  & AP$_{L}$  \\
    \toprule[1pt]
    YOLOv2 \cite{redmon2017yolo9000} & DarkNet-19 & 21.6 & 44.0  & 19.2  & 5.0 & 22.4 & 35.5  \\
    SSD512 \cite{liu2016ssd}   & VGG-16 & 28.8 & 48.5  & 30.3  & 10.9 & 31.8 & 43.5  \\
    SSD513 \cite{fu2017dssd}   & ResNet-101 & 31.2 & 50.4  & 33.3  & 10.2 & 34.5 & 49.8   \\
    DSSD513 \cite{fu2017dssd}   & ResNet-101 & 33.2 & 53.3 & 35.2  & 13.0 & 35.4 & 51.1  \\
    STDN513 \cite{zhou2018scale} &DenseNet-169 & 31.8 &51.0& 33.6 &14.4 &36.1 &43.4  \\
    FPN-Reconfig \cite{kong2018deep} & ResNet-101& 34.6 &54.3 &37.3 &- &- &- \\
    RefineDet512 \cite{zhang2018single} &ResNet-101 & 36.4 &57.5 &39.5 &16.6 &39.9 &51.4 \\
    GHM SSD \cite{li2019gradient} & ResNeXt-101 & 41.6 &62.8 &44.2& 22.3 &45.1 &55.3 \\
    CornerNet511 \cite{law2018cornernet} & Hourglass-104 & 40.5 &56.5& 43.1 &19.4 &42.7& 53.9 \\
    M2Det800 \cite{zhao2019m2det}& VGG-16 & 41.0 &59.7 &45.0 &22.1 &46.5 &53.8 \\
    ExtremeNet \cite{zhou2019bottom} &Hourglass-104& 40.2 &55.5 &43.2& 20.4 &43.2 &53.1 \\
    CenterNet \cite{zhou2019objects} &Hourglass-104 &42.1 &61.1 &45.9 &24.1 &45.5 &52.8\\
    FSAF \cite{zhu2019feature}& ResNeXt-101 & 42.9 &63.8& 46.3 &26.6 &46.2& 52.7 \\
    CenterNet511 \cite{duan2019centernet}& Hourglass-104 & 44.9& 62.4 & 48.1 &25.6 &47.4 &{\bf57.4} \\
    \hline
    RefineDet512 \cite{zhang2018single} &ResNet-101 & 36.4 &57.5 &39.5 &16.6 &39.9 &51.4 \\
    RefineDet512  & Deformable ResNet-101 & 37.9 & 58.1  & 40.2 & 18.5 & 40.4  & 51.6 \\
    RefineDet512  & GRCNN-109 & 37.6 & 58.7  & 40.5 & 18.0 & 41.3  & 52.1 \\
    RefineDet512  & Deformable GRCNN-109 & 38.4 & 59.4  & 41.0 & 18.5 & 42.1  & 52.3 \\
    RefineDet512  & SKNet-100 & 38.9 & 59.7  & 41.2 & 18.9 & 42.7  & 53.3 \\
    RefineDet512  & SK-GRCNN-109 & 39.5 & 59.9  & 41.9 & 19.8 & 43.4  & 53.7 \\
    \hline
    RetinaNet800\cite{lin2018focal}   & ResNet-101 & 39.1 &  59.1  & 42.3  & 21.8 & 42.7 & 50.2   \\
    RetinaNet800 & Deformable ResNet-101 & 40.8  & 59.6 & 42.7  & 22.7 & 43.3  & 51.0 \\
    RetinaNet800 & GRCNN-109 & 40.3  & 60.4 & 43.2  & 22.4 & 44.0  & 51.0  \\
    RetinaNet800 & Deformable GRCNN-109 & 41.2  & 60.8 & 44.3  & 23.3 & 44.3  & 51.8 \\
    RetinaNet800 & SKNet-100 & 41.6 & 61.5 & 44.7  & 23.5 & 45.1  & 52.3 \\
    RetinaNet800 & SK-GRCNN-109 & 42.3  & 61.8 & 45.4  & 23.9 & 46.0  & 52.9\\
    \hline
    FreeAnchor \cite{zhang2019freeanchor} & ResNet-101 & 43.1& 62.2 &46.4 &24.5 &46.1& 54.8  \\
    FreeAnchor & Deformable ResNet-101 & 43.5 & 64.0 &47.1  & 25.5 &  46.9 &55.7  \\
    FreeAnchor& GRCNN-109 & 44.3 & 64.5 &47.6 & 26.0 &  47.6 &56.4  \\
    FreeAnchor & Deformable GRCNN-109 & 44.8 &64.5 &47.9 &26.7 & 48.5 &56.8 \\
    FreeAnchor & SKNet-100 & 45.2 & 64.8 &48.4 & 27.5 &  48.3 & 56.6 \\
    FreeAnchor & SK-GRCNN-109 & \bf{45.6} &\bf{65.5} &\bf{48.7} &{\bf27.9} &  {\bf48.8} & 57.0 \\
     \bottomrule[1pt]
  \end{tabular}\\
  \begin{flushleft}
  \it
  The AP, AP$_{50}$ and AP$_{75}$  indicate the average precision for IoU $\in \{0.50:0.05:0.95\}$, at 0.5 and at 0.75, respectively. The AP$_{S}$, AP$_{M}$ and AP$_{L}$ denote the average precision for small sized, medium sized and large sized objects defined in \cite{lin2014microsoft} in terms of IoU $\in \{0.50:0.05:0.95\}$.
  \end{flushleft}
\end{table*}

\subsection{Pipelines}\label{sec:pipeline_Det}
To further evaluate GRCNN, we selected three popular frameworks, RefineDet \cite{zhang2018single}, RetinaNet \cite{lin2018focal}, and FreeAnchor \cite{zhang2019freeanchor}, which are categorized as ``anchor-based one-stage detectors''. The RefineDet consists of two inter-connected modules named Anchor Refinement Module (ARM) and Object Detection Module (ODM). The RetinaNet  consists of a backbone network responsible for computing feature maps over an entire input image and two task-specific networks. The FreeAnchor introduces a detection customized likelihood function into the loss function to supervise the training of RetinaNet. More details about these frameworks can be found in their original papers. One common property of these frameworks is that all of them have a backbone network (usually a pretrained ResNet-101) for feature extraction. We were interested in whether the performance would be boosted by substituting GRCNN based models with their original backbone networks.

\subsection{Implementation Details}
In the publicly released codes of the three frameworks, we replaced the backbone network ResNet-101 with the GRCNN-109 and the SK-GRCNN-109 respectively, whose network configurations are both described in Table~\ref{tab:arch-imagenet}. In addition, we also evaluated Deformable GRCNN-109 on the three frameworks. All GRCNN based models were pre-trained on the ImageNet classification task as described in Section \ref{sec:implem_imagenet}. Then it was fine-tuned using the SGD algorithm with weight decay 0.0005 and momentum 0.9. The multi-scale testing strategy was not used. The training scale range of the FreeAnchor \cite{zhang2019freeanchor} framework was set to \{640, 672, 704, 736, 768, 800\}.
The other training hyper-parameters were set according to the original papers of the three frameworks \cite{zhang2018single, zhang2019freeanchor, lin2018focal}.

\subsection{Comparison with Existing Methods} 
Table~\ref{tab:detect} presents the results of the three frameworks, as well as some other anchor-based one-stage models which do not perform multi-scale testing. 
First, we compared Deformable ResNet-101 and GRCNN-109 with ResNet-101, which were trained as described in Section~\ref{sec:results_imagenet}. The GRCNN and Deformable ConvNet obtained similar performance gains on the detection task. 
Second, the three Deformable GRCNN-109-based object detection frameworks consistently performed better than the GRCNN-109-based and Deformable ResNet-101-based counterparts, indicating that the proposed gating mechanism and Deformable convolutions are complementary with respect to improving the detection performance. 
Third, by using the SKConv, the performance of the GRCNN-109 was further improved. Finally, the SK-GRCNN based FreeAnchor achieved the best results among the anchor-based one-stage detectors compared in this study. 

\section{Concluding Remarks}\label{sec:conclusion}
The proposed model in this work is built on a previous model called recurrent convolutional neural network (RCNN) \cite{Liang2015Recurrent, Ming2015Recurrent}, which incorporates recurrent connections between neurons in the same layer. Our main idea is to introduce gates on the recurrent connections in RCNN, and the resulting model is called gated RCNN (GRCNN). The gates control the amount of information flow from neighboring neurons into each neuron according to the context. The unrelated context information coming from the recurrent connections is inhibited by the gates. Extensive experiments were carried out on some benchmark datasets for object recognition, scene text recognition and object detection. The results showed that GRCNN outperformed RCNN on all these datasets. The results verified the effectiveness of the proposed gating function. In addition, on these tasks, the proposed GRCNN often attained accuracy on par with (and sometimes, e.g., on scene text recognition, even slightly better than) the state-of-the-art models.

It is well-known that a deep neural network with more parameters has better capability to model the input-output mapping. To compete with existing models, we had to use more parameters. This was done by removing the weight sharing constraint of recurrent networks. Then the recurrent connections became conventional feedforward connections. Note that these connections determine the RFs of neurons in the resulting feedforward model, and the gates operating on these connections modulate the RFs of neurons. The effect turned out to be very good in terms of accuracy.

If accuracy is not the main concern, one does not need to break the tied weights, thus significantly reduces the number of parameters, which is favored in applications where the memory of the equipment is limited. More interestingly, one can explore to what extent the neurons in this recurrent model resemble biological neurons by measuring their non-classical RFs  (e.g., \cite{nelson1978orientation, jones2002spatial, cavanaugh2002nature}).  If the properties of artificial neurons and biological neurons do not match well, one can develop new ideas based on the proposed model to make them match better. This may lead to more brain-like and powerful computer vision models. It may also shed light on the computational mechanism of visual neurons in the brain.


%

\ifCLASSOPTIONcompsoc
  \section*{Acknowledgments}
\else
  \section*{Acknowledgment}
\fi

This work was supported in part by the National Natural Science Foundation of China under Grant Nos. 61621136008 and 61836014.

\ifCLASSOPTIONcaptionsoff
  \newpage
\fi




\bibliographystyle{IEEEtran}
\bibliography{egbib}

%

%


%
\begin{IEEEbiography}
    [{\includegraphics[width=1in,height=1.25in,clip,keepaspectratio]{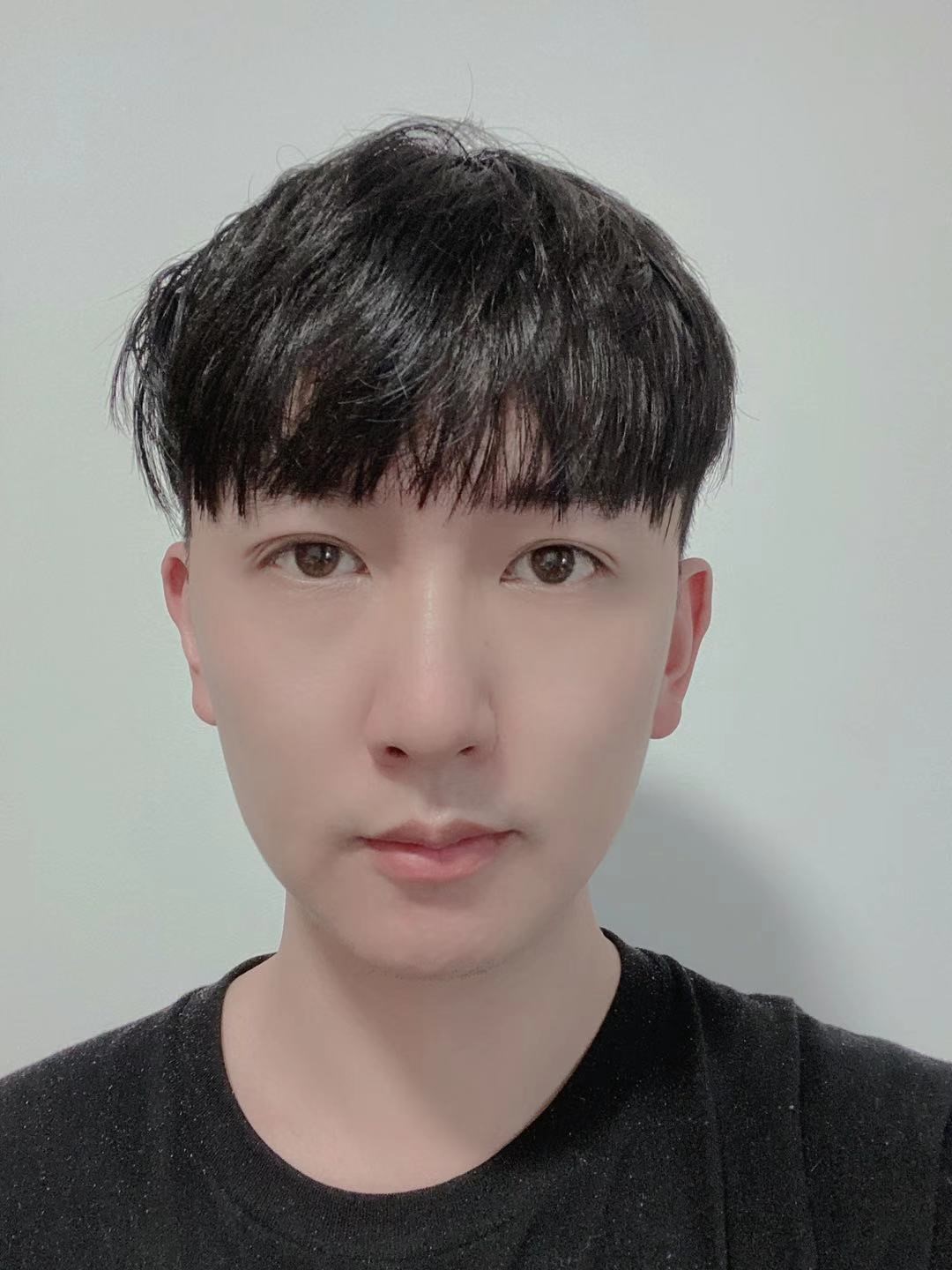}}]
    {Jianfeng Wang} received the B.E. and M.E. degrees from Beijing University of Posts and Telecommunications, Beijing, China, in 2013 and 2016 respectively. Now, he is working toward the PhD degree in the Computer Science Department from University of Oxford. His current research interests include computer vision, machine learning and deep learning.
\end{IEEEbiography}
\vspace{0cm}
\begin{IEEEbiography}
	[{\includegraphics[width=1in,height=1.25in,clip,keepaspectratio]{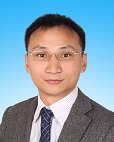}}]
    {Xiaolin Hu}  (S'01, M'08, SM'13) received the B.E. and M.E. degrees in Automotive Engineering from Wuhan University of Technology, Wuhan, China, and the Ph.D. degree in Automation and Computer-Aided Engineering from The Chinese University of Hong Kong, Hong Kong, China, in 2001, 2004, 2007, respectively. He is now an Associate Professor at the Department of Computer Science and Technology, Tsinghua University, Beijing, China. His current research interests include deep learning and computational neuroscience. He was an Associate Editor of the IEEE Transactions on Neural Networks and Learning Systems. Now he is an Associate Editor of IEEE Transactions on Image Processing and an Associate Editor of Cognitive Neurodynamics.
\end{IEEEbiography}







\end{document}